\documentclass{article}

\usepackage[preprint]{neurips_2021} %

\usepackage[utf8]{inputenc} %
\usepackage[T1]{fontenc}    %
\usepackage{url}            %
\usepackage{booktabs}       %
\usepackage{amsfonts}       %
\usepackage{nicefrac}       %
\usepackage{microtype}      %
\usepackage{xcolor}         %

\usepackage{amsmath}
\usepackage{amssymb}
\usepackage{bbm}

\makeatletter
\renewcommand\hyper@natlinkbreak[2]{#1}
\makeatother

\setcitestyle{notesep={; },round,aysep={,},yysep={;}}

\usepackage{tikz}
\usetikzlibrary{shapes,arrows}
\usetikzlibrary{backgrounds}

\usepackage{enumerate }

\usepackage{floatrow}
\newfloatcommand{capbtabbox}{table}[][\FBwidth]

\title{Bounded logit attention:\\Learning to explain image classifiers}

\author{%
  Thomas Baumhauer $\quad\quad$ Djordje Slijepcevic $\quad\quad$ Matthias Zeppelzauer\\ %
  St. P\"olten University of Applied Sciences\\
  St. P\"olten, Austria\\
  \texttt{\{firstname.lastname\}@fhstp.ac.at}
}

\begin{document}

\maketitle

\begin{abstract}
    Explainable artificial intelligence is the attempt to elucidate the workings of
    systems too complex to be directly accessible to human cognition through suitable side-information referred to as ``explanations''.
	We present a trainable explanation module for convolutional image
	classifiers we call bounded logit attention (BLA). The BLA module learns to select
	a subset of the convolutional feature map for each input instance, which then serves as
	an explanation for the classifier's prediction.
    BLA overcomes several limitations of the instancewise feature selection method ``learning to explain'' (L2X) introduced by \citet{chen2018learning}: 1) BLA scales
	to real-world sized image classification problems, and 2) BLA offers a canonical way to learn explanations of variable size.
	Due to its modularity BLA lends itself to transfer learning setups and can also be employed
	as a post-hoc add-on to trained classifiers.
	Beyond explainability, BLA may serve as a general purpose method for differentiable approximation of subset selection.
	In a user study we find that BLA explanations are preferred over explanations generated by the popular (Grad-)CAM method \citep{zhou2016learning,selvaraju2017grad}.
\end{abstract}

\section{Introduction}
\label{sec:introduction}

A commonly quoted rule of thumb is that as machine learning systems increase in size and sophistication, it becomes
increasingly hard to understand how these systems arrive at their predictions.
That is, an increase of performance is paid for by a lack of interpretability.
Uninterpretable, ``blackbox'' systems are undesirable for a multitude of reasons, both from a usability and engineering perspective.
An engineer struggling to understand a system may also struggle to diagnose and improve it
and a user who does not understand a system may (and perhaps should) not trust it.

In this work, we deal with the explanation of predictions made by image classification models based on deep convolutional neural networks.
Much work in this area is focused on generating explanations {\em post-hoc}.
This means that after building and training a model, an additional mechanism is put in place to produce explanations for the model's behavior.
An explanation could be any additional human-interpretable information suitable to improve the understanding of the workings
of (some aspects of) a model.
In the field of image classification such information could be visual, e.g. highlighting or
masking certain parts of an input image.

Considerably less popular than the use of post-hoc methods is the incorporation of trainable explanation mechanisms directly into a model. 
A drawback of this approach is that it cannot a priori be ruled out that such mechanisms could negatively affect the model's performance.
However, building  explainability mechanisms directly into a model as part of its architecture is appealing insofar as once trained such explanations provide
by construction true, accurate insight into the workings of (some aspects of) the model. This is in contrast to post-hoc explanations, where one always
has to question how faithful these explanations are to actual calculations performed by the model.

{\bf Contributions.}
We propose {\em bounded logit attention} (BLA), a trainable, modular explanation mechanism to be incorporated into convolutional image classification networks.
The BLA module learns to select
a subset of the convolutional feature map for each input instance, which then serves as
an explanation for the classifier's prediction.
Relative to previous work on learned explanations by \citet{chen2018learning}
our key contributions are:
{\bf 1)}~BLA scales to real-world sized convolutional neural networks.
{\bf 2)}~BLA offers a canonical way to produce explanations of variable size.
One of the merits of our method is that due to its modularity it can be used with typical transfer learning setups in computer vision, in particular pretrained feature extractors. Similarly, our method may even be employed as a purely post-hoc method.
Beyond explainability, BLA can be used as a general purpose method for differentiable approximation of subset selection.
In our experiments we obtain favorable results according to both quantitative metrics and human evaluation.

\section{Related work}

There is no consensus on the definitions of notions such as {\em interpretability, explainability}, etc. and we use them loosely throughout
this work. Roughly by {\em interpretable} we mean ``accessible to human understanding'' and and by {\em explanation} we mean ``any piece of
information that facilitates interpretability''. \citet{lipton2018mythos} investigates notions related to interpretability in a principled way.
In his taxonomy, the opposite of ``blackbox-ness'' is {\em transparency}. In this sense, a built-in explanation mechanism like ours makes
a model (partially) transparent.

{\bf Explanations in computer vision. }
Linear models
are easy to interpret, as long as they use a relatively small number of (interpretable) features.
As a consequence, many approaches to interpretability involve the construction of a linear surrogate model of some sort.
End-to-end differentiable models
can be linearized around an input.
In the field of computer vision this approach is known as {\em saliency maps} \citep{simonyan2013deep} which present
gradient information
graphically.
Since the gradients of large image classification networks are noisy,
averaging gradients over some neighborhood has been proposed by \citet{sundararajan2017axiomatic, smilkov2017smoothgrad}.

Another idea is decomposing input images into coarser, more interpretable features, such as superpixels (cohesive segments of similar pixels).
Then, a neighborhood of this input is defined, consisting of all images with any of the interpretable features
either ``present'' or ``absent'' (corresponding to occluding the ``absent'' superpixels with black color).
LIME \citep{ribeiro2016should} builds a local approximation of the model for this neighborhood.
Similarly, one may use the Shapley value~\citep{shapley1953value} of interpretable features
when constructing explanations.
It turns out that Shapley values also fit into the LIME framework~\citep{charnes1988extremal} as pointed out by \cite{lundberg2017unified} 
(there called SHAP).

\citet{zhou2016learning} propose {\em class activation mappings} (CAM) which assigns saliency to convolutional
feature maps based on the coefficients of a final dense linear layer. \citet{selvaraju2017grad} generalize
this approach to Grad-CAM where in the case that more than one layer is used to process the convolutional features
they use a gradient-based linear approximation of this calculation.
Example-based methods \citep{koh2017understanding, chen2019looks, hase2019interpretable}
provide explanations that relate predictions to the training data.
User studies evaluating some of the explanation methods described in this section were conducted by \citet{hase2020evaluating,jeyakumar2020can}.

{\bf Learning to explain.}
\label{sec:l2x}
Our work is inspired by the {\em learning to explain} (L2X) method by \cite{chen2018learning}, proposing ``instancewise feature selection as
a methodology for model interpretation''. Given a
$d$-dimensional input $x$ first a binary mask $\delta \in 2^d$ is computed in an explanation network (e.g. a multilayer perceptron).
Then, the masked input $x \odot \delta$ is used as the input to a second network solving the task. 
In order to make this setup end-to-end differentiable the computation of the discrete mask $\delta$ has to be approximated in a continuous fashion.
To this end, the explanation network learns a distribution of inclusion probabilities of each input feature.
Then, $k$ features are drawn independently from this distribution, to approximate (coarsely) the sampling
of a mask of size $k$ (i.e. a mask with exactly $k$ of the $d$ entries of $\delta$ equal to 1). 
The discrete sampling itself is approximated by the {\em Gumbel-softmax trick} \citep{jang2016categorical, maddison2016concrete}.
At test time no sampling is performed and instead the $k$ features with highest inclusion probabilities are used.
As a result, L2X is constrained to produce explanations of fixed size $k$, with $k$ a hyperparameter.
In this work, we propose an alternative way to compute masks that permits masks of different sizes for different inputs.
This is crucial for vision tasks as the sizes of regions of interest in an image vary between images.

For image classification \citet{chen2018learning} compute the mask $\delta$ through a convolutional network. 
As a consequence
of pooling operations in the network, $\delta$ is of lower resolution than the input $x$ and hence must be upsampled
(with repetition) to $\tilde \delta$ before computing the masked input $x \odot \tilde \delta$. The masked input image
is then classified by another convolutional network.
We find that this approach scales  poorly to real-life sized image classification problems 
(Section \ref{sec:comparision}). Furthermore, it is computationally expensive, requiring two forward passes through convolution stacks.
Our proposed architecture is designed to alleviate these issues.

For further related work and the embedding of our approach into related approaches see Section~\ref{sec:alternative}.

\section{Bounded logit attention}
\label{sec:method}

{\bf Image classification.} We consider a standard supervised image classification task for 
some dataset consisting of pairs $(x, y)$, with
$x$ an image annotated by some class-label $y$. A model $h$ for this tasks
predicts labels $\hat y = h(x)$.
We assume the following standard architecture for $h$.
Given an input image $x$ a convolutional feature extractor $\mathcal F$ computes
a list of feature vectors $\mathcal F(x) = \vec f =  \langle f_i : i < n \rangle$,
with $n$ the size of the feature map.
To avoid clutter, we use a single index $i < n$ for these features instead of the usual
two indices indicating a feature's position in the two-dimensional feature map, but of course each feature still corresponds to a region in $x$.
These features are then globally average-pooled to a feature vector
$$
v = \frac{1}{n} \sum_{i < n} f_i.
$$
Finally, a classifier $\mathcal L$ (e.g. a logistic regression head, or perhaps a multilayer perceptron)
computes
$\hat y = \mathcal L(v)$.

{\bf Explanation module.}
The features $f_i$ learned by the model $h$ may be considered an abstract representation of properties of the
corresponding regions of the input image $x$. In this work, we  are interested in selecting features that are ``important'' to the model
$h$ when predicting the label $\hat y = h(x)$. To this end, we propose to employ an explanation module
$\mathcal E$ that outputs a subset of feature indices
$\mathcal E(\vec f) = \vec \delta = \langle \delta_i : i < n \rangle \in 2^n$.
A model $h$, as described above, may then be augmented by $\mathcal E$ by replacing the average-pooling operation
by the reweighed pooling operation
$$
v = \frac{1}{\|\vec \delta\|_1} \sum_{i < n} \delta_i f_i.
$$
Then $\vec \delta = \mathcal E(\vec f)$ acts as an explanation for the prediction $\hat y = \mathcal L(v)$.
Indeed $\vec \delta$ is easily human-interpretable as each entry can be understood
as a binary flag encoding the use of the corresponding feature in the prediction,
and the size of $\vec \delta$ is typically small enough, e.g. $n = 49 = 7^2$.
Our goal is to build the explanation module $\mathcal E$ in a way such that the model remains end-to-end
differentiable, i.e. such that $\mathcal E$ may be learned.

\subsection{Parameterizing the explanation module}
\label{sec:parameterizing}

{\bf Variational approximation.}
We identify $\vec \delta = \mathcal E (\vec f)$ with
the probability distribution $p(i | \vec f) = \frac{\delta_i}{\|\vec \delta\|_1}$. 
During training we approximate $p(\cdot | \vec f)$ by some $q(\cdot | \vec f) =
\mathcal Q(\vec f)$ where $\mathcal Q$ is an element of some variational family
$$
V = \{\mathcal Q_\phi \mid \mathcal Q_\phi : \vec f \mapsto q(\cdot | \vec f), \phi \in \Phi
\}
$$
parameterized by $\phi$. Hence, we need to find a suitable variational family, which in deep learning
terms means finding the appropriate neural architecture implementing $\mathcal Q$. 
We require $\mathcal Q$ to compute a probability distribution $q$ approximating
$p$.

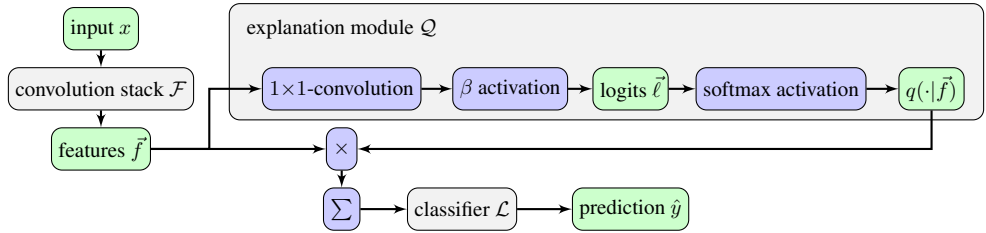
\begin{figure}[t]
	\vskip .5em
	
	\centering
	\tikzstyle{block} = [rectangle, draw, fill=blue!20, 
	text centered, rounded corners, minimum height=2em]
	\tikzstyle{line} = [draw, -latex']
	
	\tikzstyle{block2} = [rectangle, draw, fill=green!20,
	text centered, rounded corners, minimum height=2em] %
	\tikzstyle{block3} = [rectangle, draw, fill=black!5,
	text centered, rounded corners, minimum height=2em] %
	
	\begin{tikzpicture}[node distance = 1cm, auto, scale=0.8, every node/.style={transform shape}]
	
	\node [block2] (input) {input $x$};
	\node [block3,below of=input] (conv) {convolution stack $\mathcal F$};
	\node [block2,below of=conv] (features) {features $\vec f$};
	
	\node [block,right of=conv, node distance = 4cm] (onetimesone) {$1{\times}1$-convolution};
	\node [block,right of=onetimesone, node distance = 2.8cm] (beta) {$\beta$ activation};
	\node [block2,right of=beta, node distance = 2cm] (logits) {logits $\vec \ell$};
	\node [block,right of=logits, node distance = 2.5cm] (softmax) {softmax activation};
	\node [block2,right of=softmax, node distance = 2.5cm] (q) {$q(\cdot | \vec f)$};

	\node [block,right of=features, node distance = 4cm] (product) {$\times$};
	\node [block,below of=product] (sum) {$\sum$};
	\node [block3,right of=sum, node distance = 2cm] (lr) {classifier $\mathcal L$};

	\node [block2,right of=lr, node distance = 2.8cm] (prediction) {prediction $\hat y$};

	\node [above of=onetimesone] (x-module) {explanation module $\mathcal Q$};
	
	\begin{scope}[on background layer]
	
	\node [block3, minimum height=5.5em, text width=35em] at (8.4,-0.55) {};
	\end{scope}

	\path [line, thick] (input) -- (conv);
	\path [line, thick] (conv) -- (features);
	\path [line, thick] (features.east) -- (1.8,-2) -- (1.8,-1) -- (onetimesone.west);
	\path [line, thick] (onetimesone) -- (beta);
	\path [line, thick] (beta) -- (logits);
	\path [line, thick] (logits) -- (softmax);
	\path [line, thick] (softmax) -- (q);
	\path [line, thick] (q.south) |- (product);
	\path [line, thick] (features) -- (product);
	\path [line, thick] (product) -- (sum);
	\path [line, thick] (sum) -- (lr);
	\path [line, thick] (lr) -- (prediction);
	
	\end{tikzpicture}
	
	\caption{Image classifier augmented with a bounded logit attention explanation module.}
	\label{fig:x-module}
	
\end{figure}

{\bf Activation function.} The key trick we propose is using the simple, yet uncommon activation function
$$
\beta(x) = \min(x, 0)
$$
in the approximation of the logits of such discrete distributions. The $\beta$-activation function is of course 
closely related to the popular ReLU function. What is uncommon is applying it directly before the softmax function.

Concretely, we build $\mathcal Q$ as follows (Figure~\ref{fig:x-module}). Given a feature map $\vec f$, we apply a $1{\times}1$-convolution with
a single filter, mapping each $f_i$ to a scalar value $g_i = u^\top f_i$, with $u$ the weights
of the convolution.\footnote{
	Of course more complicated mappings $f_i \mapsto g_i$ could be considered, however we found
	a single $1{\times}1$-convolution to be sufficient.
} Then, we compute the logits
$\vec \ell = \langle \ell_i : i < n \rangle$ where $\ell_i = \beta(g_i)$. Finally, we apply the softmax function $\sigma$
to obtain $\vec q = \sigma(\theta \cdot \vec \ell)$, with the temperature~$\theta$ being a hyperparameter.
During training we plug $\mathcal Q$ into the original model $h$ by replacing the pooling operation with
\begin{align}
	\label{form:pooling}
	v = \sum_{i < n} q(i | \vec f) f_i.
\end{align}

{\bf Discretization. }After training we obtain the explanation module $\mathcal E$ from its variational approximation~$\mathcal Q$ 
by defining $\delta_i = 1$ iff $\ell_i = 0$. The merit of the $\beta$-function now becomes clear:
it allows for constructing explanations of variable size (the size of an explanation being $\|\vec \delta\|_1$) by giving us a canonical way
of discretization. Importantly, it forces $q$ to be uniform on indices for which $\ell_i = 0$, making the
behavior of $\mathcal Q$ and its discretization $\mathcal E$ match more closely.

{\bf Thresholding.} Why may we hope that $\mathcal Q$ might learn to approximate reasonable explanations? Remember that we are interested in finding ``important'' features. 
The explanation module $\mathcal Q$ computes weights
$q(i|\vec f)$ for each feature $f_i$ to use in the pooling operation in formula (\ref{form:pooling}).
Thus during training $\mathcal Q$ should learn to assign more weight to features that are discriminative for the task, and to mostly ignore features that are not.
To encourage this behavior and put additional learning pressure on $\mathcal Q$ we propose to threshold weights
in a modified pooling operation
\begin{align}
	v = \sum_{i < n} \mathbbm{1}[q(i | \vec f) > \gamma] f_i
\end{align}
where $\mathbbm{1}[q(i | \vec f) > \gamma]$ is $1$ if $q(i | \vec f) > \gamma$
and $0$ otherwise.
We propose $\gamma = \frac{1}{n}$ as the default threshold.

\subsection{Discussion}
\label{sec:method-discussion}

{\bf Bounded logit attention (BLA).} One way to think about the explanation module described above is to consider it an
{\em attention mechanism}, in the sense that $\mathcal Q$
computes soft attention, approximating hard attention in $\mathcal E$. With the $\beta$-activation function
bounding the logits of $q(\cdot | \vec f)$ from above being the key property of this attention mechanism, we refer to our architecture as
{\em bounded logit attention} (BLA). Independently of explainability,
BLA can serve as a general purpose method to produce differentiable approximations
for the selection of subsets.

{\bf Transfer learning.} Note that the modularity of our method lends itself to transfer learning setups. Having trained a model $h = \mathcal F \circ \mathcal L$, we
may for example hold both $\mathcal F, \mathcal L$ fixed and only
train an explanation module $\mathcal Q$ as a post-hoc addition to $h$.
We investigate this approach in Section~\ref{sec:main-experiment} under the name 
BLA-PH. In this setup we could even make predictions using the original model $h$
(without $\mathcal Q$ plugged in) and use the outputs of $\mathcal Q$ as a purely post-hoc explanation.  Section~\ref{sec:faithfulness} provides justification for this
approach (that has the advantage that the outputs of $\mathcal Q$
are far cheaper to compute than the LIME scores used there).
Finally, our method can readily be used with a pretrained feature extractor $\mathcal F$
while training both $\mathcal L, \mathcal Q$,
an approach we take throughout Section~\ref{sec:main-experiment}.

\begin{figure}[t]
    \begin{floatrow}
    \ffigbox[5.5in]{%
	\begin{center}
	    \scalebox{.8}{
	    {\small Cats vs. dogs $\quad\quad\quad\quad\quad\quad$
	    Stanford dogs $\quad\quad\quad\quad\quad\quad$
	    Caltech birds
	    $\quad\quad\quad\quad\quad\quad\quad\quad\quad\quad\quad\quad\quad\quad\quad\quad$ }}\\
	     \scalebox{.8}{
	    \rotatebox{90}{\small $~~~$ Vis. $~~~~~~$
	    Hard $~~~~~~$
	    Soft $~~~~~~$
	    Input}
	    }
	    \scalebox{.8}{
		\includegraphics[width=0.8\linewidth]{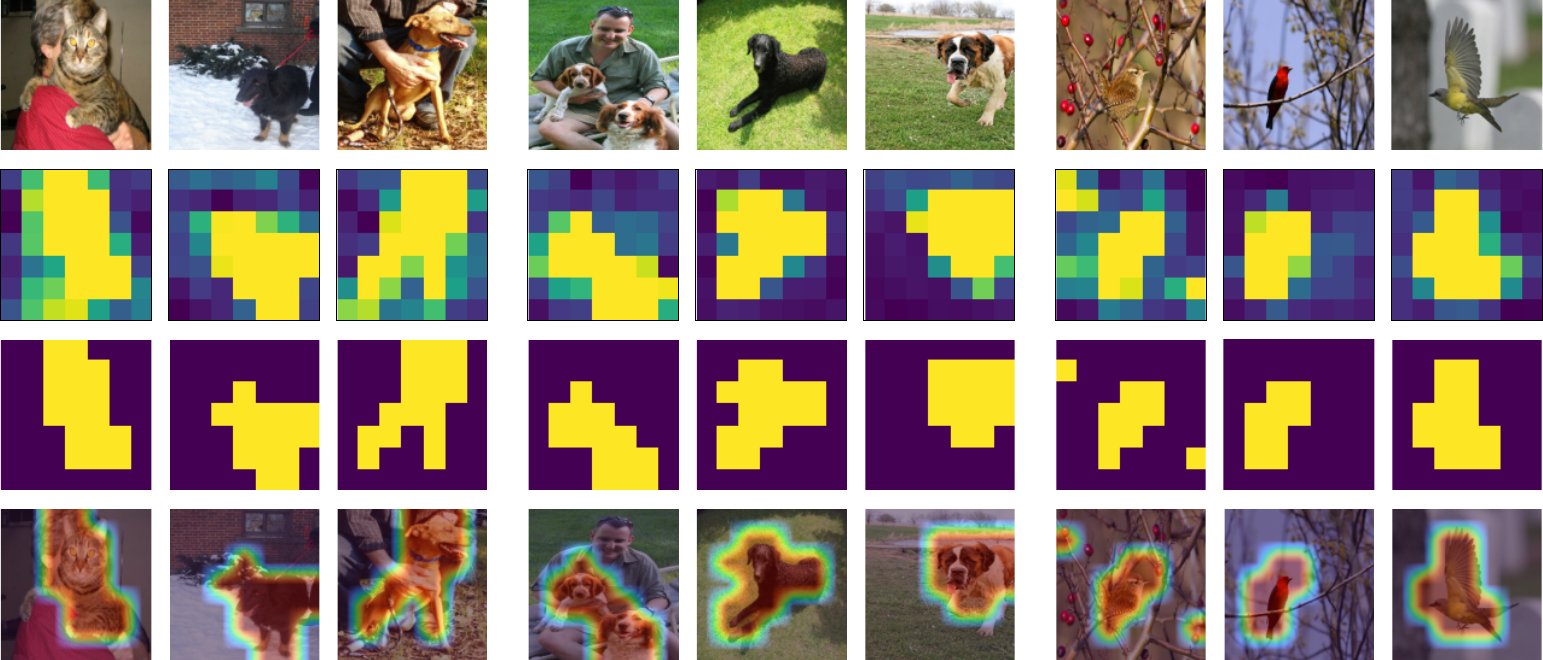}
		}
		$\quad$
		\includegraphics[height=3.85cm]{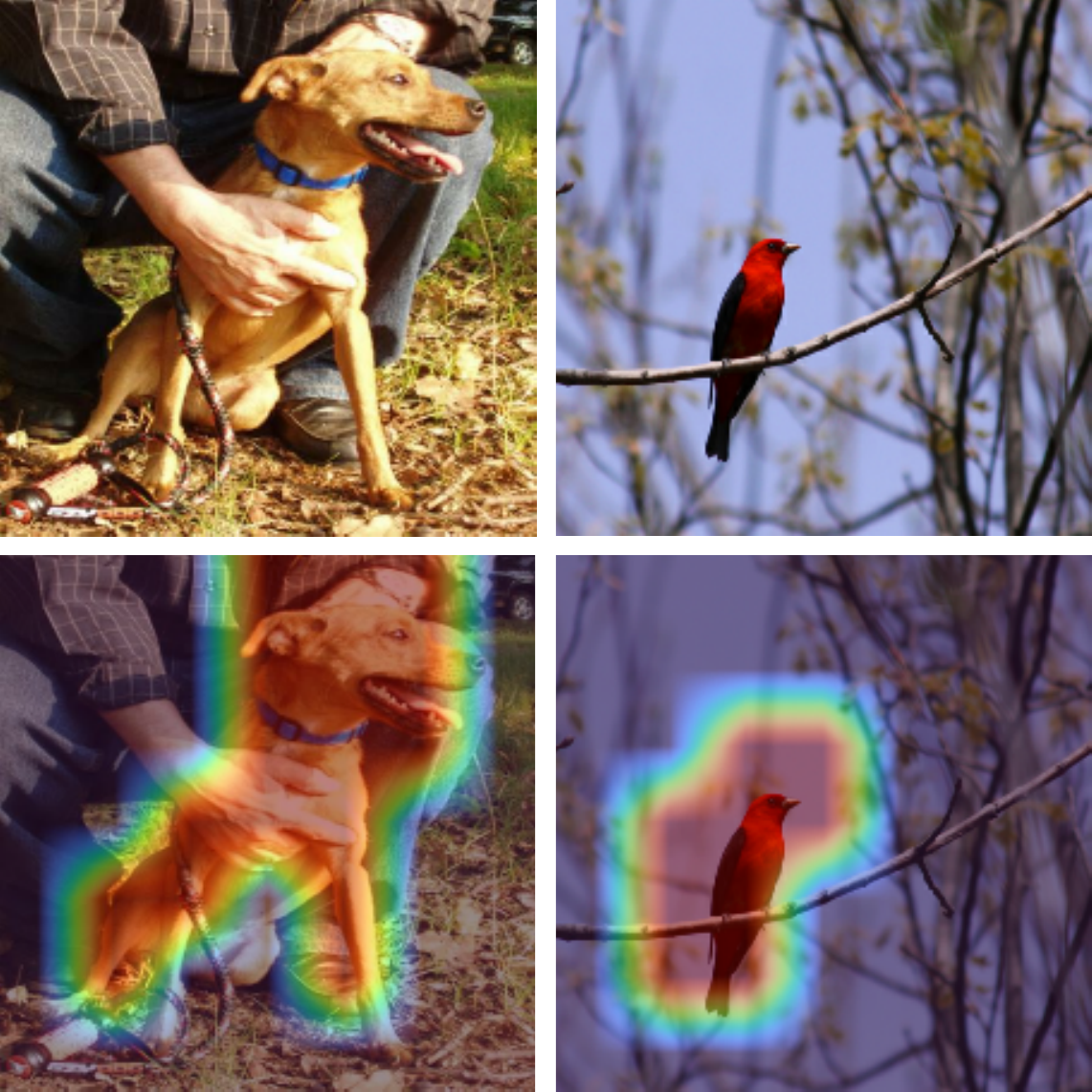}
	\end{center}
	}{%
	\vskip -.5em
	\caption{Left: Learned explanations by BLA 
	on three datasets from the experiment in Section~\ref{sec:main-experiment}.
	Rows, from top to bottom: input image;
	soft explanation $q(i | \vec f)$ computed by  $\mathcal Q$;
	discretized hard test-time explanation $\vec \delta$
	by $\mathcal E$;
	visualization of $\vec \delta$ as described in Section~\ref{sec:method-discussion}.
	Right: enlarged version.}
	\label{fig:examples}
	}
	\end{floatrow}
	\vskip -1.2em
\end{figure}

{\bf Visualization.} Like the (Grad-)CAM method \citep{zhou2016learning,selvaraju2017grad}, BLA produces explanations at the level of the convolutional feature map.
In Section~\ref{sec:introduction} we mentioned that a conceptual advantage of the learned explanations produced by~$\mathcal E$ is that they faithfully reflect the workings of (some aspect of) the model by construction.
However, we still need to make a choice how to present the outputs~$\vec \delta$ of~$\mathcal E$.
Following the convention of the (Grad-)CAM literature, we visualize
$\vec \delta$ using a colormap, which is upscaled and imposed on the
input image (Figure \ref{fig:examples}).
The experiment in Section~\ref{sec:faithfulness} provides justification for our choice.
(Other choice are possible of course, e.g. to spread out each $\delta_i$ over
the corresponding receptive field.)
Note that both (Grad-)CAM and BLA achieve interpretability
by restricting their semantics to  the most easily interpretable aspect of convolutional features which is their location.

\subsection{Alternative explanation modules}
\label{sec:alternative}

In Section~\ref{sec:parameterizing} we proposed BLA as a possible architecture of the explanation module  $\mathcal Q$.
There exist several concepts in the literature one might consider for alternative explanation modules.

{\bf Fixed size explanations: L2X-F.}
When employing the transfer learning approach discussed above,
keeping the feature extractor $\mathcal F$ fixed, and identifying each input $x$ with its feature map $\vec f = \mathcal F(x)$,
our approach resembles computing L2X (see Section~\ref{sec:l2x}) explanations for the classification head $\mathcal L$.
The key differences are: L2X a priori fixes the size of the generated explanation to some $k < n$,
while in our approach use of the $\beta$-function allows to generate explanations of a variable size.
Like BLA, L2X computes a probability distribution over the input points (which here are the convolutional feature vectors), but unlike 
BLA the training of L2X has a non-deterministic element in the approximate sampling from this distribution.
We refer to L2X at the feature level as {\em L2X-F} and use it as a baseline to compare our approach to.

Another way to think about the BLA explanation module is that $q(\cdot|f_i)$ computes
an {\em objectness score} for the feature $f_i$.
However, the prediction of the objectness  of $f_i$ is not made independently of the objectness of $f_j, j \neq i$.
As a result of the softmax function there is global competition
between these scores, in the sense that increasing $q(\cdot|f_i)$ necessarily means decreasing $\sum_{j \neq i} q(\cdot|f_j)$.
Thus the softmax function provides a weak form of global context in the explanation module.
Is this the right amount of context?

{\bf More context: attention with global concept vector.}
\citet{jetley2018learn} propose ``an end-to-end differentiable attention module for  convolutional neural network architectures built for image classification''.
Like in our approach, the authors compute probability distributions to weight features (and they do this for three different
levels of convolutional features, not just the outputs of the feature extractor as we do).
The key differences to our approach is that a global concept vector $g$, computed as a learned linear transformation of
the concatenation of all feature vectors is used in the computation of the logits of the probability distribution.
To this end, $g$ is added to each feature vector $f_i$, resulting in logits $\ell_i = u^\top (g + f_i)$.
We attempted incorporating a global concept vector in the BLA module in this way (while keeping the the $\beta$-activation for the logits before
the softmax). Interestingly, this resulted in the following peculiar 
behavior. The module learned to compute extremely negative logits (of magnitude ca. $-1000$), essentially circumventing
the $\beta$-activation function (Figure \ref{fig:global-concept} in the appendix). As a result, only a single feature is chosen in the discrete attention module $\mathcal E$
(since there is now a unique maximum of the logits $\ell_i$).
We believe this is due to the concept vector enabling the explanation module to ``cheat'' in the sense that it can be
very certain which feature vector it wants to choose.
Since the initial results were not promising, we did not pursue this approach further.

{\bf Less context: pointwise attention.}
\citet{park2018bam} and \citet{woo2018cbam} propose an attention module for convolutional neural networks they call CBAM.
They do not use the softmax function when computing their attention maps, but instead use pointwise sigmoid activations, i.e.
they use no global context at all.
We found that for our purposes the lack of competition between features
did not seem to put sufficient learning pressure on an explanation module to choose between features, resulting in poorly focused attention maps.
Again, we did not pursue this approach further.

\section{Experiments}

\subsection{Comparison to L2X}

\label{sec:comparision}
In this section we substantiate the claims that 1) BLA
improves accuracy over L2X on small datasets and 2) unlike BLA, L2X scales poorly
to larger datasets, hence a new method is indeed needed.

The MNIST dataset \citep{lecun2010mnist} contains $28{\times}28$ grayscale images of handwritten digits. We subsample all images of 3 and 8, resulting in 11982 training and 1984 validation images.
\citet{chen2018learning} report validation accuracy of 0.958 (using hard explanations, there called post-hoc accuracy) on this dataset for their L2X method, with fixed explanation size $k=4$.
For their baseline model without learned explanations they report a validation accuracy of 0.997.
We imitate their CNN architecture, building a model consisting of three
$5{\times}5$ convolutions with ReLU activations with 8, 16, and 16 filters, respectively. The first two
convolutions are followed by $2{\times}2$ maximum pooling, and the third convolution
is followed by a dense linear layer with a sigmoid output unit. We augment this CNN by a BLA explanation
module as described in Section~\ref{sec:parameterizing}, and train the model end-to-end for 3 epochs, using the Adam optimizer \citep{kingma2014adam} with a learning rate of $10^{-3}$ to minimize the cross-entropy loss.
We use temperature $\theta = 0.1$ and thresholding with $\gamma = 49^{-1}$.
Over 20 runs we obtain a mean baseline accuracy of
$0.994 \pm$1.53e-3, and a mean accuracy of $0.993 \pm$3.86e-4 for the interpretable model, with no statistically significant difference (p=0.26 Mann-Whitney-U-test).

The cats vs. dogs dataset \citep{asirra-a-captcha} consists of 23,262 medium resolution color images,
divided equally into two classes ``cats'' and ``dogs''. We resize all images to $224{\times}224$. We use 18,609
images for training and 4653 images for validation.
We adapt the setup of \cite{chen2018learning} to the cats vs. dogs dataset.
The explainer network uses an EfficientNet-B0  \citep{tan2019efficientnet} convolution stack (pretrained on ImageNet and frozen in the subsequent experiments), producing explanations of size $k=8$ (8 selected $32{\times}32$ patches of the $224{\times}224$ input image), while the classification network uses the same convolution stack, followed by a dense, linear layer with sigmoid activation. We train this setup for 3 epochs using the Adam optimizer
\citep{kingma2014adam}
with a learning rate of $10^{-3}$. L2X obtains a validation accuracy of $0.96$ using
the soft explanations,
but a validation accuracy of $0.5$ when using the desired  hard explanations (i.e. the random baseline). According to our visual assessment, the produced explanations indeed failed to capture meaningful information.
We take the failure of this experiment as evidence that L2X is difficult to
scale to deep CNNs (while also being less efficient,
requiring two separate forward passes through convolution stacks per prediction, while our method only requires one).

\subsection{Understanding the BLA explanation module}
\label{sec:main-experiment}

In this section we show that BLA improves accuracy and loss over the
L2X-F baseline, implying that the improvements of BLA over L2X
are not solely due to the switch to a single pass convolution architecture. We proceed to show that thresholding slightly increases accuracy further, while increasing loss.
Finally, we show that employing BLA-T post-hoc does not decrease
accuracy on average.

{\bf Datasets A,B,C.}
We denote the cats vs. dogs dataset described above by A.
Additionally, we evaluate our method on the Stanford Dogs dataset \citep{KhoslaYaoJayadevaprakashFeiFei_FGVC2011, imagenet_cvpr09}, denoted by B, consisting of 20,580 medium size color images of 120 dog breeds. We use 12,000 images for training and 8,580 for validation. Finally, the Caltech-UCSD Birds-200-2011 dataset
\citep{WelinderEtal2010}, denoted by C, consists of 11,788 color images of
200 bird species. We use 5,994 images for training and 5,794 for validation. For either dataset images are resized to $224{\times}224$.

We use EfficientNet-B0 \citep{tan2019efficientnet} as
our classification model $h = \mathcal F \circ \mathcal L$, i.e. $\mathcal F$ is the EfficientNet-B0
convolution stack and $\mathcal L$ consists of a dropout layer \citep{srivastava2014dropout}, dropping units with probability $0.2$,
followed by a dense linear layer with softmax activations.
The parameters of $\mathcal F$ are pretrained on ImageNet
and held fixed
in our experiments.
Throughout this section we use the Adam optimizer \citep{kingma2014adam}
with a learning rate of~$10^{-3}$, a batch size of 32, minimizing the cross-entropy loss.
Non-pretrained parameters are set using the Glorot uniform initalizer \citep{glorot2010understanding}.

{\bf Baseline (BL).}
We begin by training classification heads $\mathcal L$ for each dataset
(training for 2 epochs for A and 5 epochs for B, C), to obtain trained classification models $h = \mathcal F \circ \mathcal L$.
These models will serve as the uninterpretable baseline. Note that our goal is
not to reach state-of-the-art performance, but to investigate how the addition of explainability modules affects the performance,
relative to the baseline.
We conduct every experiment in this section 20 times for each dataset, e.g.
we begin by training 20 baseline models for each dataset. To compare the results we then use the Wilcoxon signed-rank test, setting the level of significance to 0.05. When talking about measurements (e.g. accuracy or loss) we always refer to the mean over all 20 runs. When comparing mean measurements, their differences are always implied to be statistically significant, unless stated otherwise.

\begin{table}[t]
	\begin{center}
		\caption{Accuracy and loss for datasets A,B,C, for the experiments in Section~\ref{sec:main-experiment}.
			The symbol $\dagger$ denotes no statistically significant difference to the baseline (BL) column.
			The symbol $*$ denotes the same for the column to the left. The value after the $\pm$ symbol
			is the standard error multiplied by 100.}
		\label{tab:measurements}
		\rotatebox{90}{\footnotesize$\mkern-60mu$ Loss $~~$ Accuracy}
		\begin{tabular}{c c c c c c } 
			 & BL & L2X-F & BLA & BLA-T & BLA-PH \\ 
			\hline
			A & 0.9912 $\pm$ 0.01 & 0.9767 $\pm$ 0.12 & 0.9902 $\pm$ 0.02 & 0.9905$^{*}$ $\pm$ 0.01 & 0.9902$^{*}$ $\pm$ 0.02 \\
			B & 0.8357 $\pm$ 0.03 & 0.7730 $\pm$ 0.24 & 0.8234 $\pm$ 0.07 & 0.8269 $\pm$ 0.03 & 0.8265$^{*}$ $\pm$ 0.07 \\
			C & 0.6794 $\pm$ 0.07 & 0.6451 $\pm$ 0.12 & 0.6722 $\pm$ 0.09 & 0.6801$^{\dagger}$ $\pm$ 0.06 & 0.6650$^{\dagger *}$ $\pm$ 0.62 \\
			\hline
			A & 0.0237 $\pm$ 0.03 & 0.0800 $\pm$ 0.33 & 0.0349 $\pm$ 0.09 & 0.0742 $\pm$ 0.15 & 0.0562 $\pm$ 0.40 \\
			B & 0.5313 $\pm$ 0.08 & 1.4042 $\pm$ 0.76 & 0.8357 $\pm$ 0.33 & 1.1319 $\pm$ 0.29 & 1.0649$^{*}$ $\pm$ 2.53 \\
			C & 1.2988 $\pm$ 0.09 & 2.3237 $\pm$ 1.37 & 1.7809 $\pm$ 0.59 & 2.1510 $\pm$ 1.09 & 2.1714$^{*}$ $\pm$ 1.63 \\
			\hline
		\end{tabular}
	\end{center}
\end{table}

\begin{figure}[t]
    \vskip -1em
    \begin{floatrow}
    \ffigbox[5.5in]{%
	\begin{center}
	    \rotatebox{90}{\footnotesize $~~~\quad\quad\quad$ Error rate}
		\includegraphics[height=3.1cm]{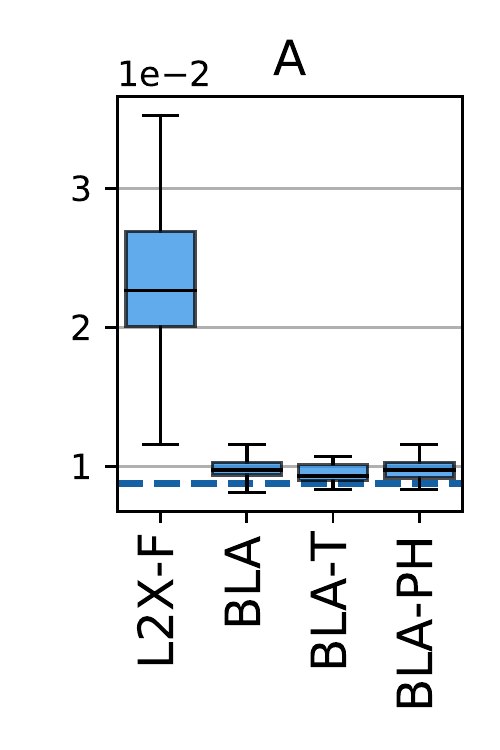}
		\includegraphics[height=3.1cm]{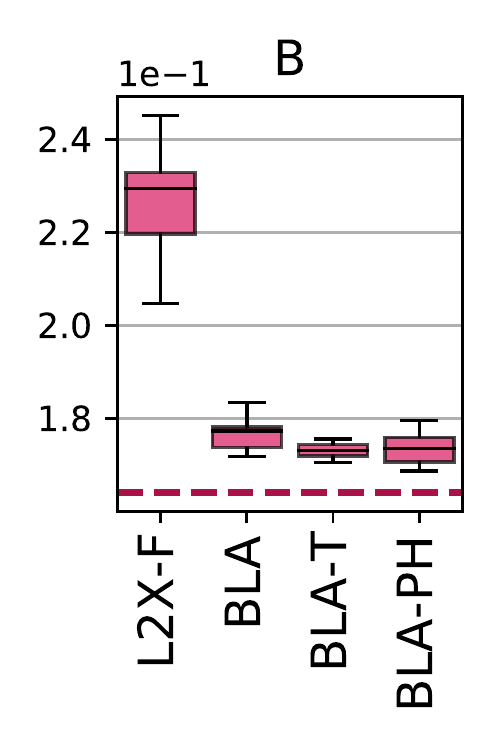}
		\includegraphics[height=3.1cm]{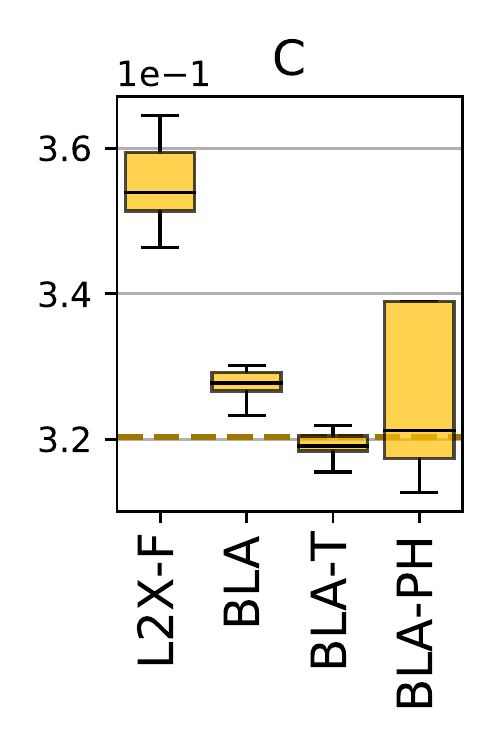}
		$\quad$
	    \rotatebox{90}{\footnotesize $~~\quad\quad\quad\quad$ Loss}
		\includegraphics[height=3.1cm]{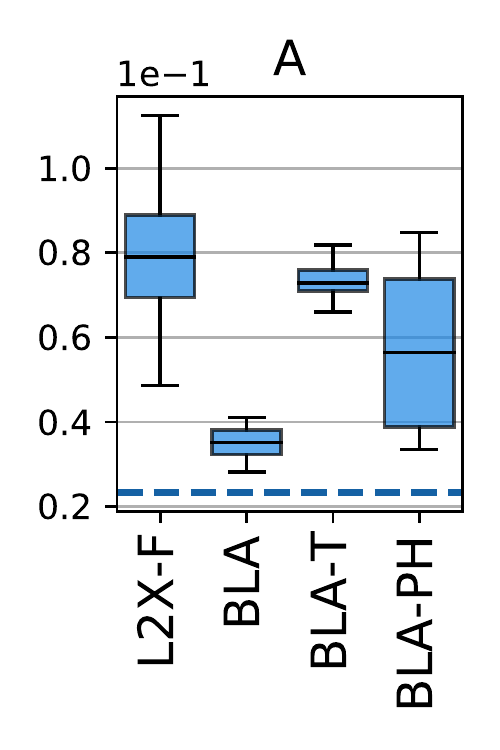}
		\includegraphics[height=3.1cm]{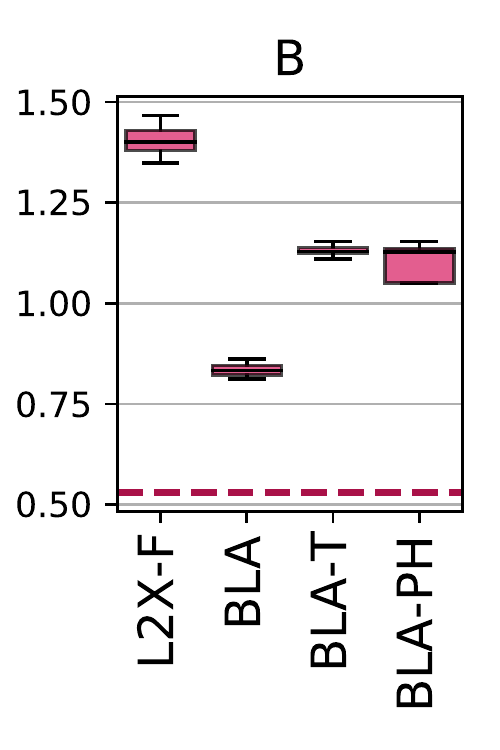}
		\includegraphics[height=3.1cm]{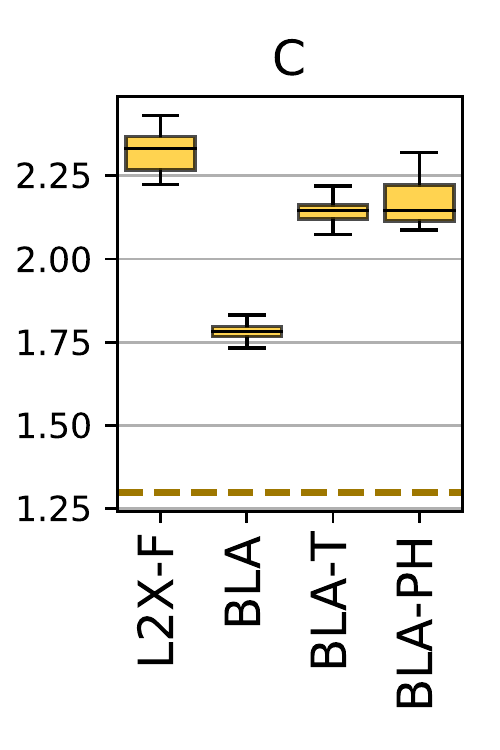}
	\end{center}
}{%
    \vskip -1.2em
	\caption{
	Graphical representation of Table~\ref{tab:measurements}.
	Left: error rate ($= 1-$accuracy), right
	loss (lower is better for either).
	The dashed line represents the median of the uninterpretable baseline models.
	We observe improvements of both error rate and loss
	for BLA (ours) over L2X-F.
	}
	\label{fig:boxplots}
	}
	\end{floatrow}

    \vskip -.8em
\end{figure}

In a next step, we augment our baseline models $h = \mathcal F \circ \mathcal L$
with explanation modules $\mathcal Q$. We train the resulting models
for 2 epochs for dataset A and 5 epochs for B,C. The parameters of $\mathcal F$ and $\mathcal L$ serve as a starting point for training, with $\mathcal Q$ randomly initialized as described above. The convolution stack $\mathcal F$ continues to be held fixed, while $\mathcal L, \mathcal Q$ are trained, except in the last
experiment where we hold the head $\mathcal L$ fixed as well.
Validation accuracies and losses are always reported for the hard explanations
generated by $\mathcal E$ (which are approximated during training by
soft explanations in $\mathcal Q$).
The numerical results of the following experiments are listed in Table~\ref{tab:measurements} and presented graphically in Figure~\ref{fig:boxplots}. Examples of  explanations for this experiment are shown in Figure~\ref{fig:examples} and further examples
from experiments in this section are found in the appendix
(Figure \ref{fig:first}-\ref{fig:last}).

{\bf L2X-F.}
First, we try adding L2X-F modules to obtain an interpretable baseline
as discussed in Section~\ref{sec:alternative}.
Accuracy and loss for all three datasets are considerably
worse than the corresponding measurements for the baseline, i.e.
adding interpretability through L2X-F modules comes at a considerable cost
in model performance.

{\bf BLA.}
Next, we investigate our BLA modules, with a temperature $\theta = 0.1$ and without thresholding.
The accuracies improve greatly compared to the L2X-F interpretable baseline,
while not quite reaching the uninterpretable baseline.
The losses also improve compared to the L2X-F baseline, but do not come quite as close
to the baseline losses.

{\bf BLA-T.}
Next, we investigate the BLA modules, again with temperature $\theta = 0.1$
using thresholding with $\gamma = 0.98 / 49 = 0.02$.
Compared to BLA (the same method without thresholding), the accuracies improved slightly. For dataset A the
improvement is not statistically significant, for B the improvement is significant with the accuracy still significantly
worse than the uninterpretable baseline, and for C the accuracy obtained is
no longer significantly different to uninterpretable baseline.
Curiously, the losses increased a fair bit compared to BLA, but remained lower than the L2X-F losses.

{\bf BLA-PH.}
Finally, we try the same setup as BLA-T, but with frozen heads $\mathcal L$,
i.e. only the explanation module $\mathcal Q$ being updated during training.
Except for the loss for dataset A, all of the
measurements are not statistically significantly different from the
corresponding measurements for BLA-T. This means models
with $\mathcal L$ frozen perform on average as well as those where $\mathcal L$ is trained
together with $\mathcal Q$. However, the variances of these measurements increased drastically.

We ran our experiments on
NVIDIA GeForce GTX 1080 GPUs. While each of the models for any of the three datasets can be trained in a matter of minutes, due the large number of runs the experiments described above take up about
11 hours of GPU time. 
Our implementation is available at: \texttt{https://github.com/th-b/bla}

{\bf ImageNet.} We demonstrate that our method scales  to the ImageNet 2010 dataset by \cite{ILSVRC15},
consisting of medium resolution color images of 1000 categories. We train on 
1,261,406 images and validate on 50,000, resizing to $224{\times}224$.
In the fashion described above, we train one baseline model, and models augmented with BLA and BLA-T modules, for 2 epochs each. We obtain an uninterpretable baseline accuracy
of 0.469. For BLA we obtain 0.458 and for BLA-T 0.525, improving upon the baseline accuracy.
An additional 8 hours of GPU time were used.

\begin{figure}[t]
\begin{floatrow}
\ffigbox{%
	\begin{center}
		\includegraphics[height=2.4cm]{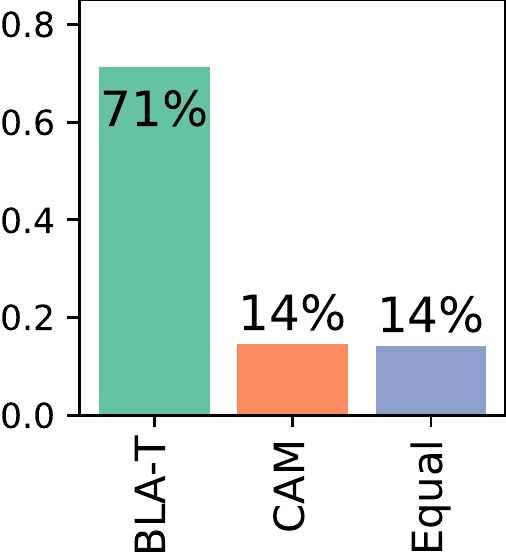}
		\includegraphics[height=2.4cm]{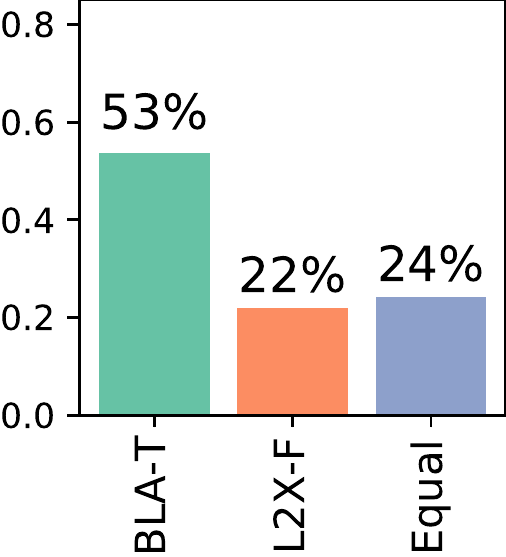}
		\includegraphics[height=2.4cm]{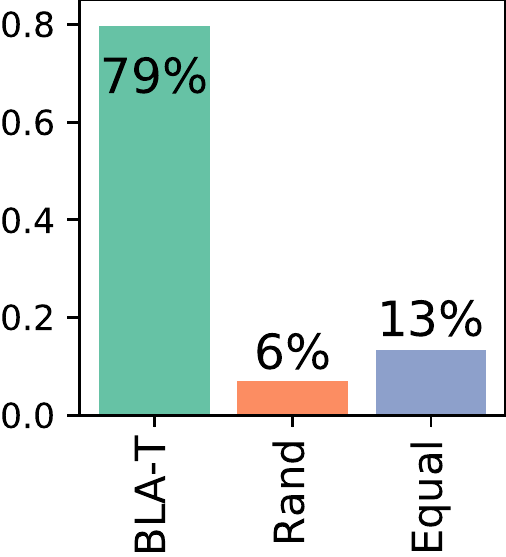} $~$
	\end{center}
	}{%
	\vskip -2em
	\caption{ User preference for explanations, from left to right: BLA-T vs. (Grad-)CAM; BLA-T vs. L2X-F; BLA-T vs. a random explanation from the same distribution.}
	\label{fig:userstudy_results}
	}
$\quad$
\capbtabbox{%
  \begin{tabular}{ccc}
  Dataset & LIME/CAM & LIME/BLA-T \\ \hline
  A & 0.47 & 0.37 \\
  B & 0.33 & 0.47 \\
  C & 0.08 & 0.37 \\ \hline
  \end{tabular}
}{%
  \caption{Mean Spearman rank correlation between
  LIME and CAM, respectively LIME and BLA-T explanations.}%
  \label{tab:corrs}
}

\end{floatrow}
\vskip -1em
\end{figure}

\subsection{On faithfulness}
\label{sec:faithfulness}

While learned explanation mechanisms are faithful to the model by construction, getting
back to points raised in Section~\ref{sec:method-discussion}
we may wonder: a) How faithful are BLA explanations to the uninterpretable baseline model, i.e. is it appropriate to use
BLA in a post-hoc setup and b) is our choice of visualization faithful?
Inspired by an occlusion experiment in \cite{selvaraju2017grad},
we use the LIME method \cite{ribeiro2016should}, which was designed for local 
faithfulness, as a baseline.
For the uninterpretable baseline model, we compute LIME scores for patches of size $32{\times}32$ of the
$224{\times}224$ input images from the validation set, resulting in explanations of size $7{\times}7$,
by randomly sampling 1000 occlusion patterns and fitting a least squares model.
We compute the Spearman rank correlation between these LIME explanations and
the $7{\times}7$ soft explanations produced by $\mathcal Q$ (using BLA-T).
Additionally we compute rank correlation between
LIME and (Grad-)CAM (Table~\ref{tab:corrs}).
We observe that both BLA-T and (Grad-)CAM consistently achieve moderate rank correlation, except on dataset C where the rank correlation for (Grad-)CAM
is low. We conclude that our method's faithfulness is comparable to that
of (Grad-)CAM on datasets A,B and considerable more faithful on dataset C.

\subsection{User study}

In a user study we investigate the following questions:
{\bf 1)} How do users rate BLA-T explanations compared to explanations generated by
the popular (Grad-)CAM method \citep{zhou2016learning,selvaraju2017grad} which also
produces explanations at the level of the convolutional feature map
and {\bf 2)} compared to L2X-F fixed size explanations?
{\bf 3)} Can users tell the difference between BLA-T and random explanations from the same distribution
(a BLA-T explanation from another, random image in the dataset imposed on the original image)?
The latter question is intended as a sanity check to verify if our method actually produces meaningful output, as it is easy to fool oneself by wishful thinking when relying on one's own visual assessment
\citep{adebayo2018sanity}.

\begin{figure}[t]

    \begin{floatrow}
    \ffigbox[5.5in]{%
	\begin{center}
	\frame{
		\includegraphics[width=.75\linewidth]{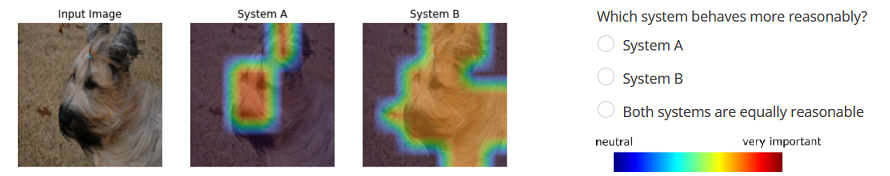}
	}
	\end{center}
	}{%
	\vskip -.5em
	\caption{
	A question to users, here L2X-F (``System A'', not to be confused with dataset A) vs.
	BLA-T (``System B''). Instructions were: ``Two systems (A and B) were trained to differentiate dog breeds. Both systems decide for the class *briard* in this example. Below, the two different systems show you on which parts of the image they base their decision. Which system behaves more reasonably?''
	}
	\label{fig:user_study_mockup}
	}
	\end{floatrow}
	\vskip -1em
\end{figure}

The participants in this study were 62 unpaid volunteers from diverse educational backgrounds.
In each question, participants were presented with a random image from one of the datasets A,B,C and two explanations, corresponding to one of the three questions above. The position of the explanations was randomized.
Participants were asked which of the two explanations, if any, they consider more reasonable.
For each question type and dataset, participants answered 4 questions (Figure~\ref{fig:user_study_mockup}).

We obtain the following results (Figure~\ref{fig:userstudy_results}):
{\bf 1)}
When compared to (Grad-)CAM, BLA-T was chosen 71\% of times and (Grad-)CAM 14\% of times.
We consider such clear user preference for our method over the well established (Grad-)CAM
method to be strong evidence that BLA will be able to serve as a useful new tool in the field of explainable
artificial intelligence.
{\bf 2)} When compared to L2X-F, BLA-T was chosen 53\% of times and L2X-F 22\% of times. The difference between these methods
is less decisive than for the previous question, as evidenced by
the 24\% ``equally reasonable'' responses, however overall there is
still a clear preference for BLA-T. Thus, our method is an improvement
over ``learning to explain'' \cite{chen2018learning} on feature level
not only conceptually by being able to produce explanations of 
variable size, but also according to user preference.
{\bf 3)} Our sanity check is passed easily, with 79\% choosing the
true BLA-T explanations and only 6\% the random explanation.
A more detailed evaluation is found in the appendix (Figures \ref{fig:user-study-appendix}-\ref{fig:user-study-appendix-last}).

\section{Conclusion}
\label{sec:conclusion}

We presented bounded logit attention (BLA) as a method to
learn explanations in deep image classification networks, selecting ``important''
elements of the convolutional feature map for each instance.
The key idea of using the activation function $\beta(x) = \min(x,0)$ to bound the logits of the attention map allowed us to canonically produce explanations of variable size.
Beyond explainability, BLA can also serve a general purpose method to produce differentiable  approximations for the selection of subsets.

The main concern with trainable, built-in explainability methods is that they
affect model performance. 
In Section~\ref{sec:comparision} we showed that unlike L2X method of \cite{chen2018learning} our method does not affect accuracy
on a subsample of the MNIST dataset. Our experiment attempting to scale L2X
to the cats vs. dogs dataset failed, demonstrating the need for a more
scalable approach such as the one presented in this work.
In Section~\ref{sec:main-experiment} we
showed that L2X-F, i.e. using the L2X method on feature level,
also comes at a considerable cost of decreased accuracy.
However, using BLA modules the performance came close to that of the
uninterpretable baseline models. The accuracies slightly improved further
when using thresholding in the BLA module, which for Caltech birds datatset even resulted in accuracy statistically indistinguishable from the baseline accuracy.
However, thresholding came at the price of increased loss.
A pure transfer learning approach, only training
$\mathcal Q$ while holding the head $\mathcal L$ fixed, resulted in no change
in performance on average, but increased variance. 

Our quantitative experimental results are thus summarized as follows:
built-in explainability for an image classification model using the BLA attention module comes at a very moderate cost of accuracy, while (unlike post-hoc methods)
providing by construction faithful insight into the workings of the classifier.
Finally, conducting a user study, we found strong evidence that participants prefer
our method over (Grad)-CAM and L2X on feature level.

\newpage

\bibliography{bounded_logit_attention}

\newpage

\section*{Checklist}

The checklist follows the references.  Please
read the checklist guidelines carefully for information on how to answer these
questions.  For each question, change the default \answerTODO{} to \answerYes{},
\answerNo{}, or \answerNA{}.  You are strongly encouraged to include a {\bf
justification to your answer}, either by referencing the appropriate section of
your paper or providing a brief inline description.  For example:
\begin{itemize}
  \item Did you include the license to the code and datasets? \answerYes{See Section~\ref{gen_inst}.}
  \item Did you include the license to the code and datasets? \answerNo{The code and the data are proprietary.}
  \item Did you include the license to the code and datasets? \answerNA{}
\end{itemize}
Please do not modify the questions and only use the provided macros for your
answers.  Note that the Checklist section does not count towards the page
limit.  In your paper, please delete this instructions block and only keep the
Checklist section heading above along with the questions/answers below.

\begin{enumerate}

\item For all authors...
\begin{enumerate}
  \item Do the main claims made in the abstract and introduction accurately reflect the paper's contributions and scope?
    \answerYes{We did our best.}
  \item Did you describe the limitations of your work?
    \answerYes{We discuss the conceptual disadvantages of built-in, learned explanations over post-hoc explanations.}
  \item Did you discuss any potential negative societal impacts of your work?
    \answerNA{One of the goals of explainable artificial intelligence 
    is the mitigation of negative societal impacts of artificial intelligence.}
  \item Have you read the ethics review guidelines and ensured that your paper conforms to them?
    \answerYes{}
\end{enumerate}

\item If you are including theoretical results...
\begin{enumerate}
  \item Did you state the full set of assumptions of all theoretical results?
    \answerNA{}
	\item Did you include complete proofs of all theoretical results?
    \answerNA{}
\end{enumerate}

\item If you ran experiments...
\begin{enumerate}
  \item Did you include the code, data, and instructions needed to reproduce the main experimental results (either in the supplemental material or as a URL)?
    \answerYes{Will be provided in the supplemental material. Git url is
    redacted for anonymity. }
  \item Did you specify all the training details (e.g., data splits, hyperparameters, how they were chosen)?
    \answerYes{}
	\item Did you report error bars (e.g., with respect to the random seed after running experiments multiple times)?
    \answerYes{}
	\item Did you include the total amount of compute and the type of resources used (e.g., type of GPUs, internal cluster, or cloud provider)?
    \answerYes{See section \ref{sec:main-experiment}}
\end{enumerate}

\item If you are using existing assets (e.g., code, data, models) or curating/releasing new assets...
\begin{enumerate}
  \item If your work uses existing assets, did you cite the creators?
    \answerYes{\cite{lecun2010mnist,asirra-a-captcha,KhoslaYaoJayadevaprakashFeiFei_FGVC2011,WelinderEtal2010}}
  \item Did you mention the license of the assets?
    \answerYes{Our code is GPLv3}
  \item Did you include any new assets either in the supplemental material or as a URL?
    \answerYes{}
  \item Did you discuss whether and how consent was obtained from people whose data you're using/curating?
    \answerYes{All participants were volunteers.}
  \item Did you discuss whether the data you are using/curating contains personally identifiable information or offensive content?
    \answerNA{}
\end{enumerate}

\item If you used crowdsourcing or conducted research with human subjects...
\begin{enumerate}
  \item Did you include the full text of instructions given to participants and screenshots, if applicable?
    \answerYes{Figure \ref{fig:user_study_mockup}}
  \item Did you describe any potential participant risks, with links to Institutional Review Board (IRB) approvals, if applicable?
    \answerNA{}
  \item Did you include the estimated hourly wage paid to participants and the total amount spent on participant compensation?
    \answerYes{unpaid}
\end{enumerate}

\end{enumerate}

\appendix

\section{User study}

\begin{figure}[h!]
	\begin{center}
		\includegraphics[width=0.7\linewidth]{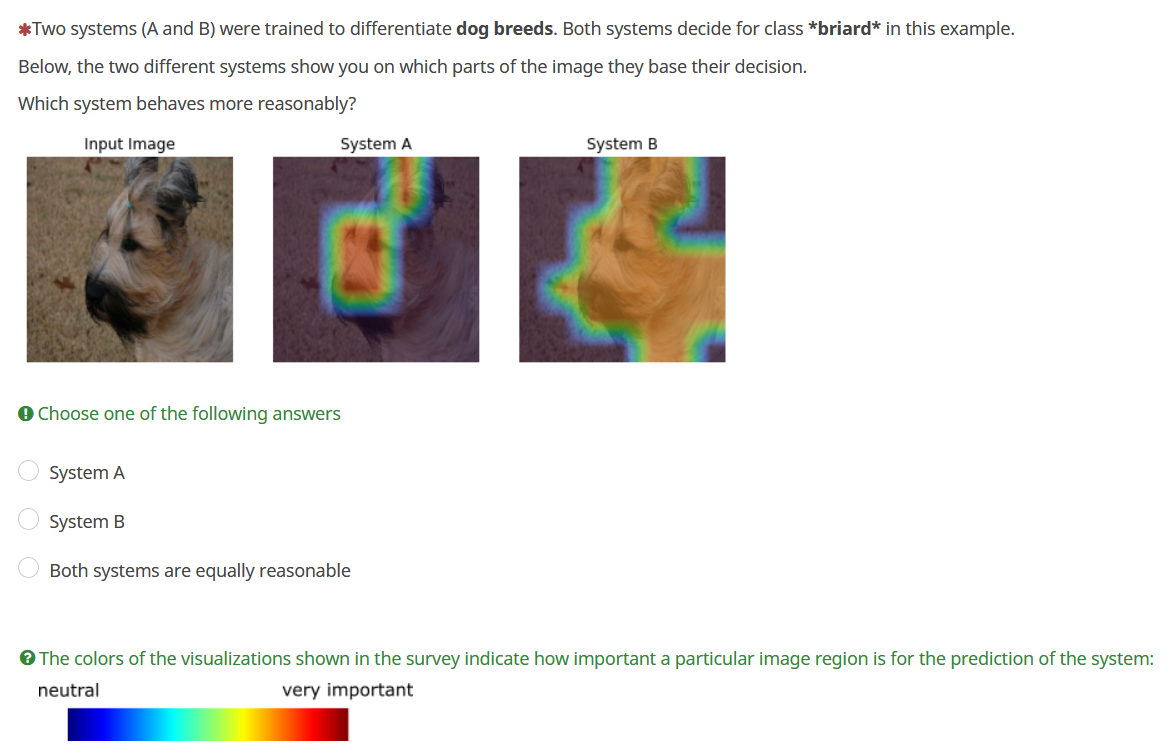}
	\end{center}
	
	\caption{A question to users, here L2X-F (``System A'', not to be confused with dataset A) vs.
	BLA-T (``System B''). Non-rearranged  version of Figure
	\ref{fig:user_study_mockup}.}
	\label{fig:user_study}
\end{figure}

\begin{figure}[h!]
	\begin{center}
	    \includegraphics[width=1\linewidth]{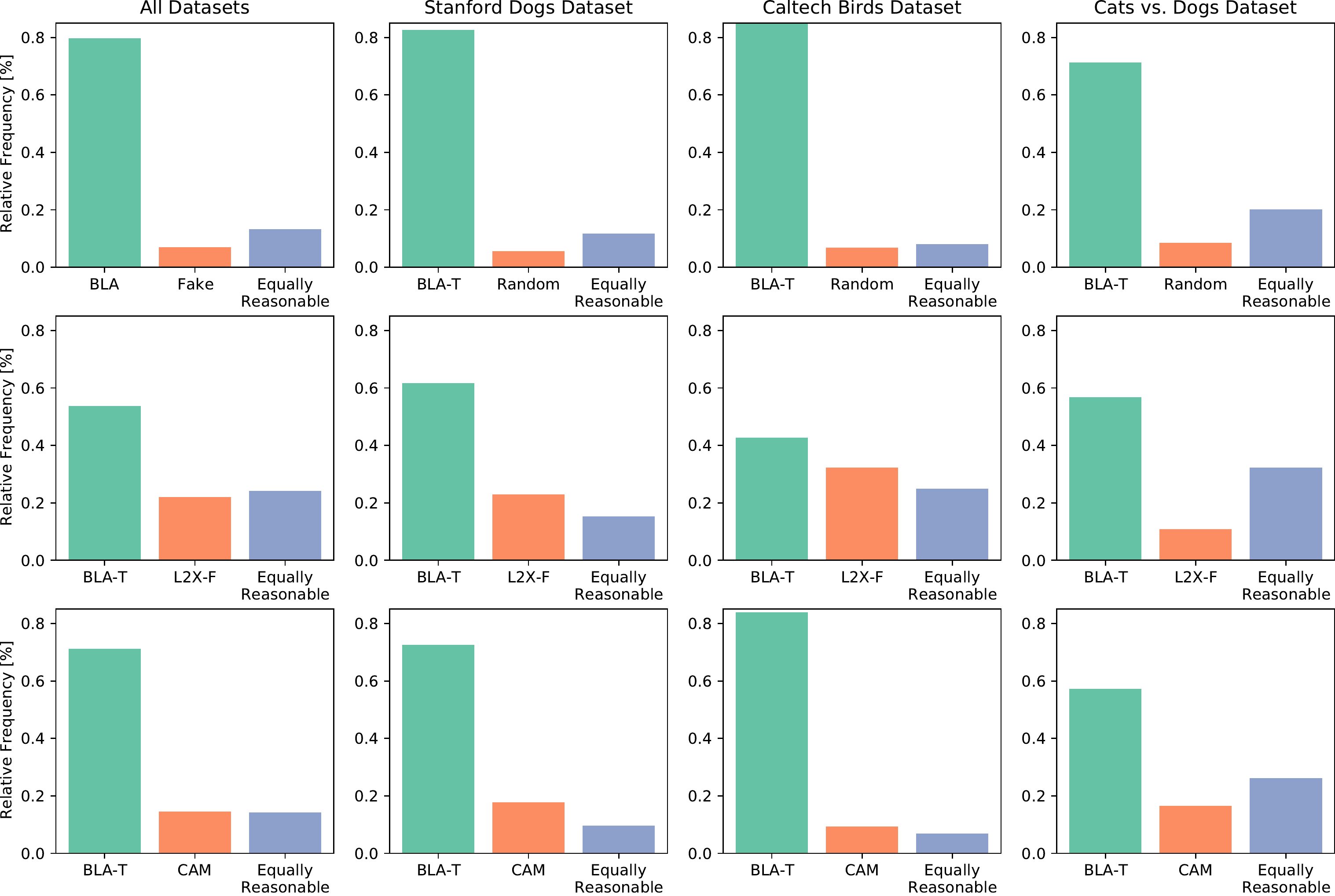}
	\end{center}
	
	\caption{More detailed results of user study: Figure \ref{fig:userstudy_results} by datasets A = Cats vs dogs, B = Stanford dogs, and C = Caltech birds.}
	\label{fig:user-study-appendix}
\end{figure}

\begin{figure}[h!]
	\begin{center}
	    \includegraphics[width=0.35\linewidth]{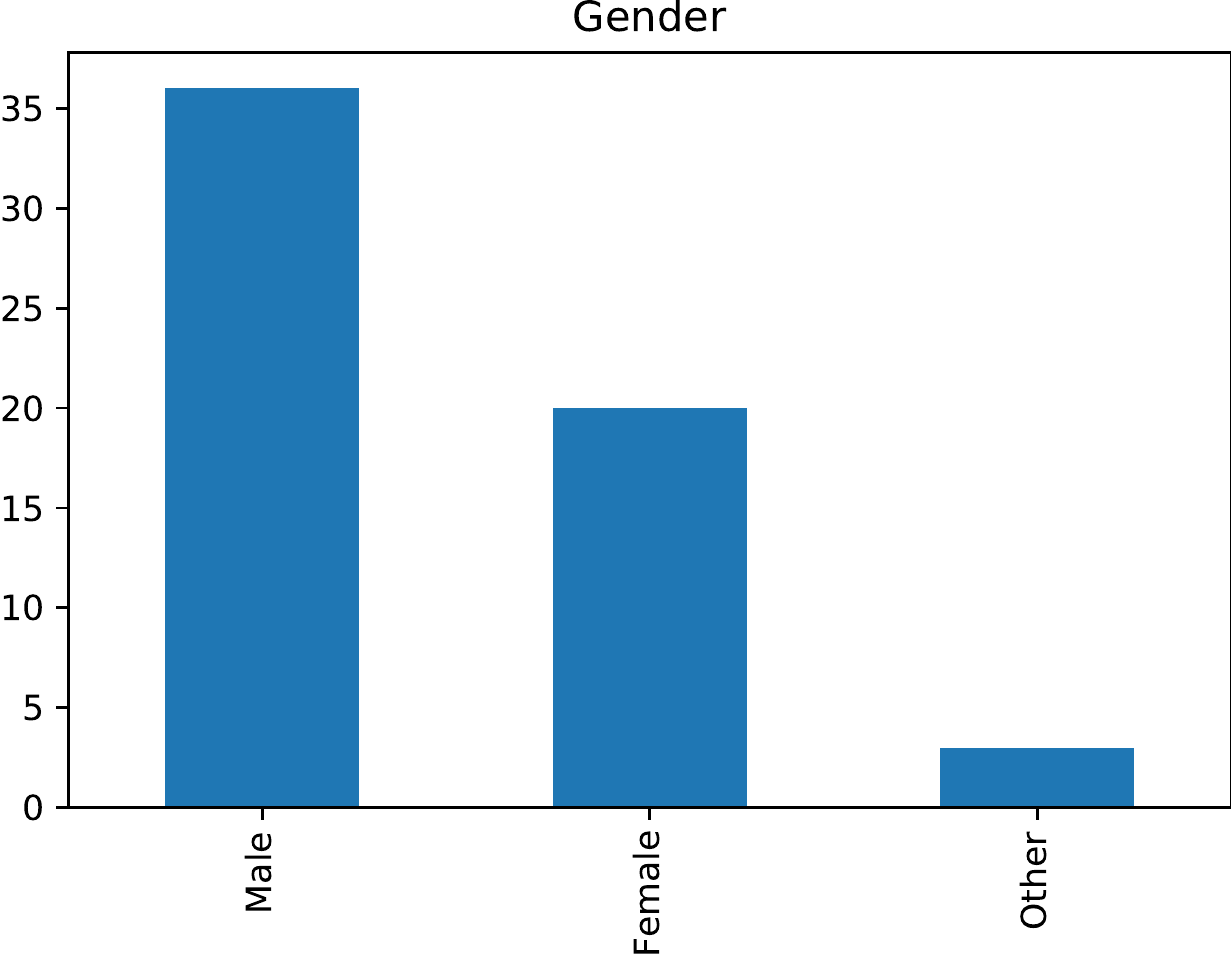}
	\end{center}
	
	\caption{Gender of the study participants.}
\end{figure}

\begin{figure}[h!]
	\begin{center}
	    \includegraphics[width=0.35\linewidth]{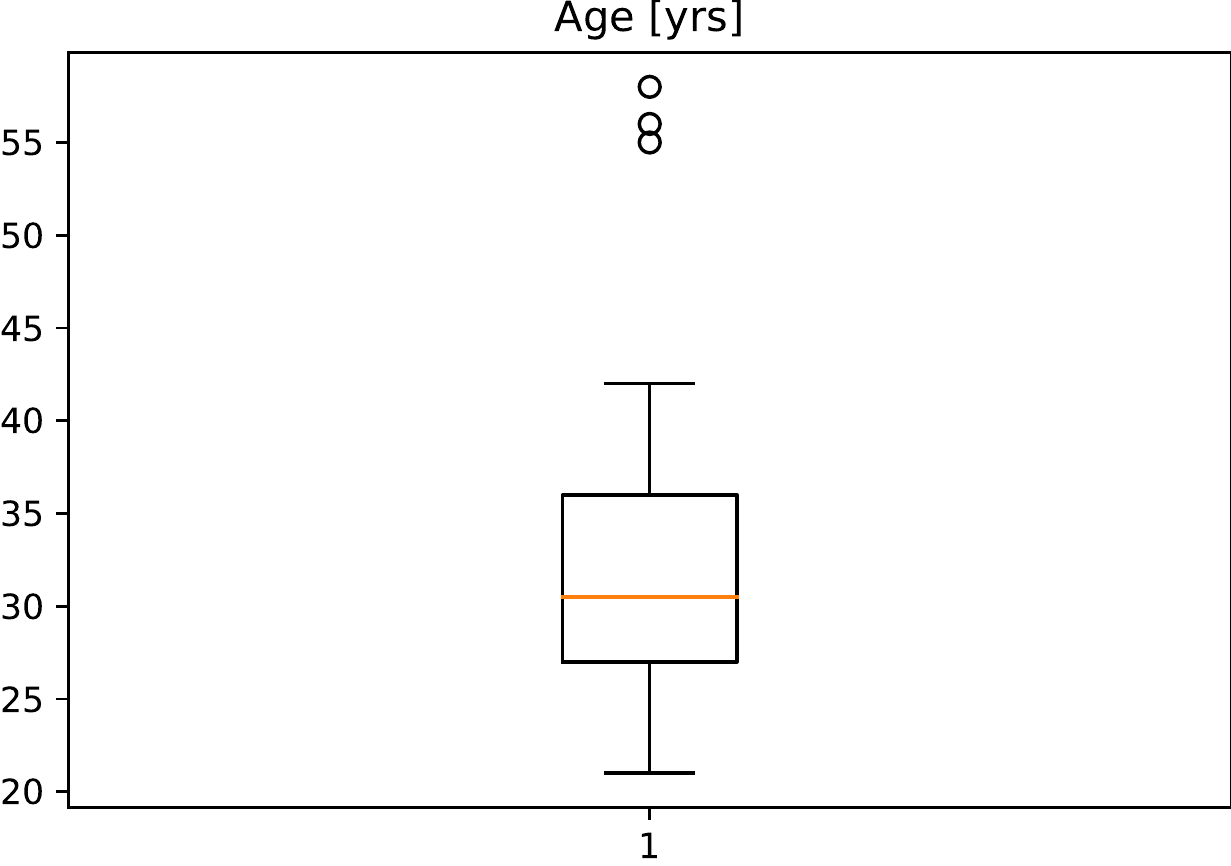}
	\end{center}
	
	\caption{Age of the study participants.}
\end{figure}

\begin{figure}[h!]
	\begin{center}
	    \includegraphics[width=0.35\linewidth]{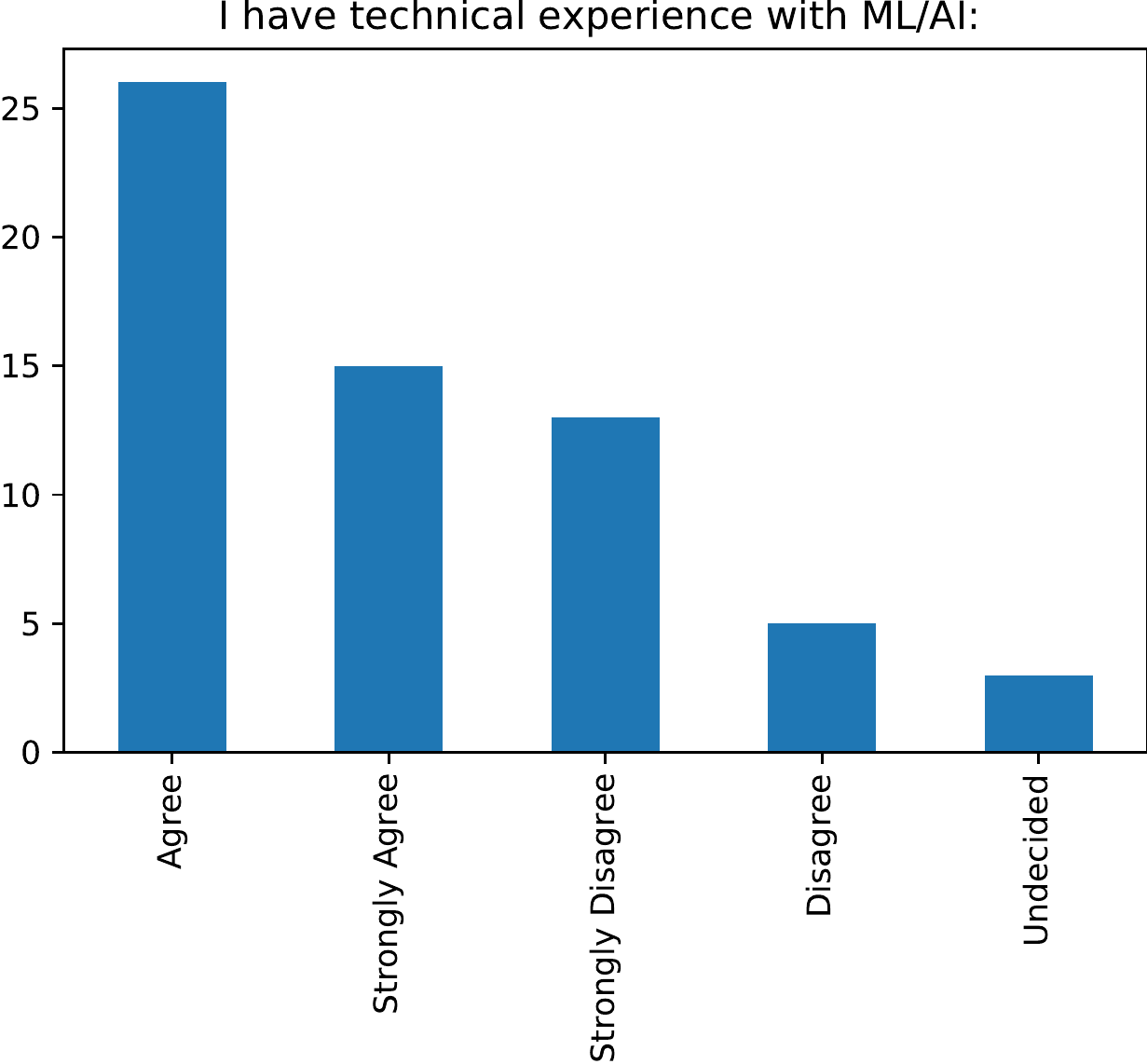}
	\end{center}
	
	\caption{Machine learning/artificial intelligence (ML/AI) experience of study participants.}
\end{figure}

\begin{figure}[h!]
	\begin{center}
	    \includegraphics[width=0.35\linewidth]{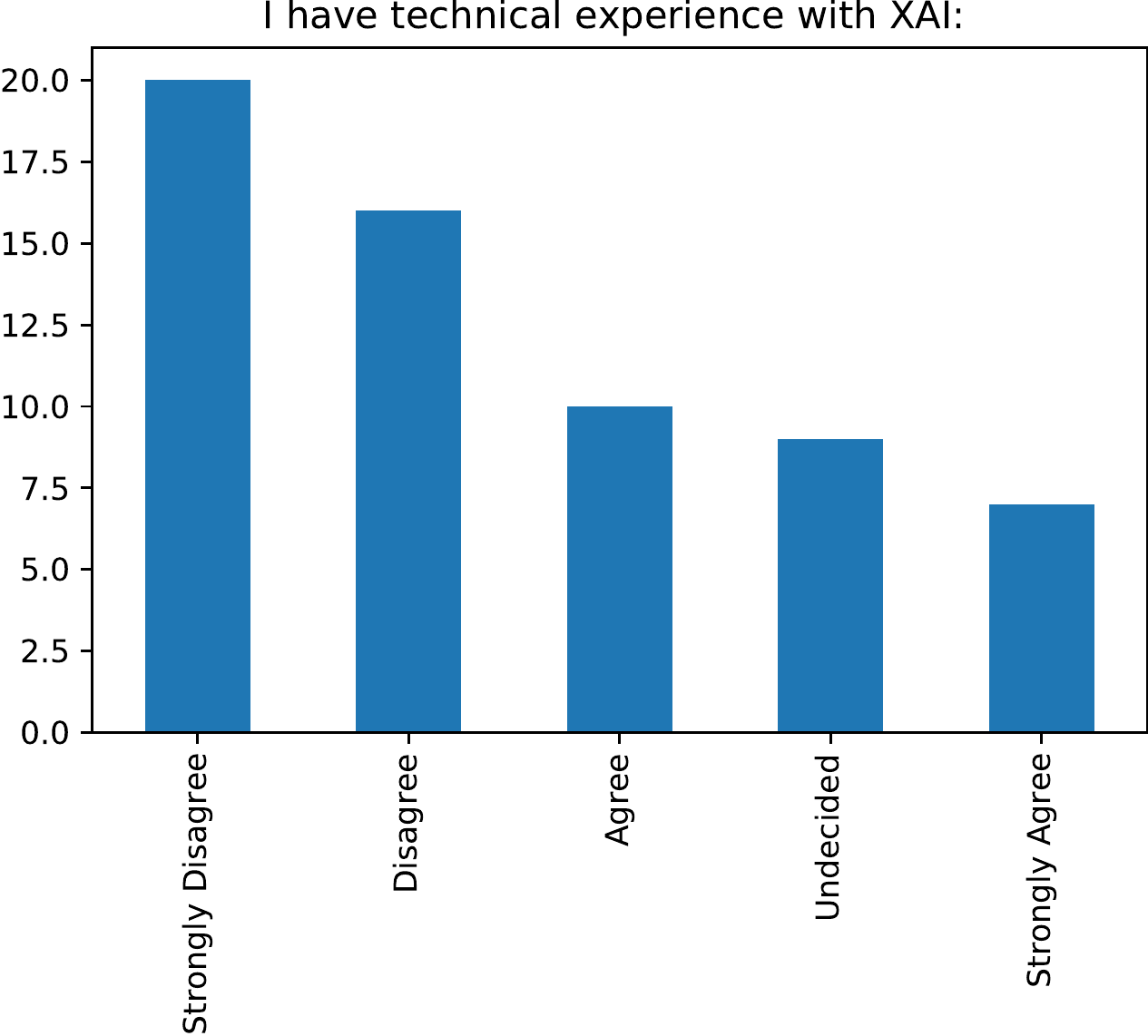}
	\end{center}
	
	\caption{Explainable artificial intelligence (XAI) experience of study participants.}
	\label{fig:user-study-appendix-last}
\end{figure}

\newpage
\clearpage
\section{Understanding the BLA explanation module}

\begin{figure}[h!]
	\begin{center}
		\includegraphics[width=0.5\linewidth]{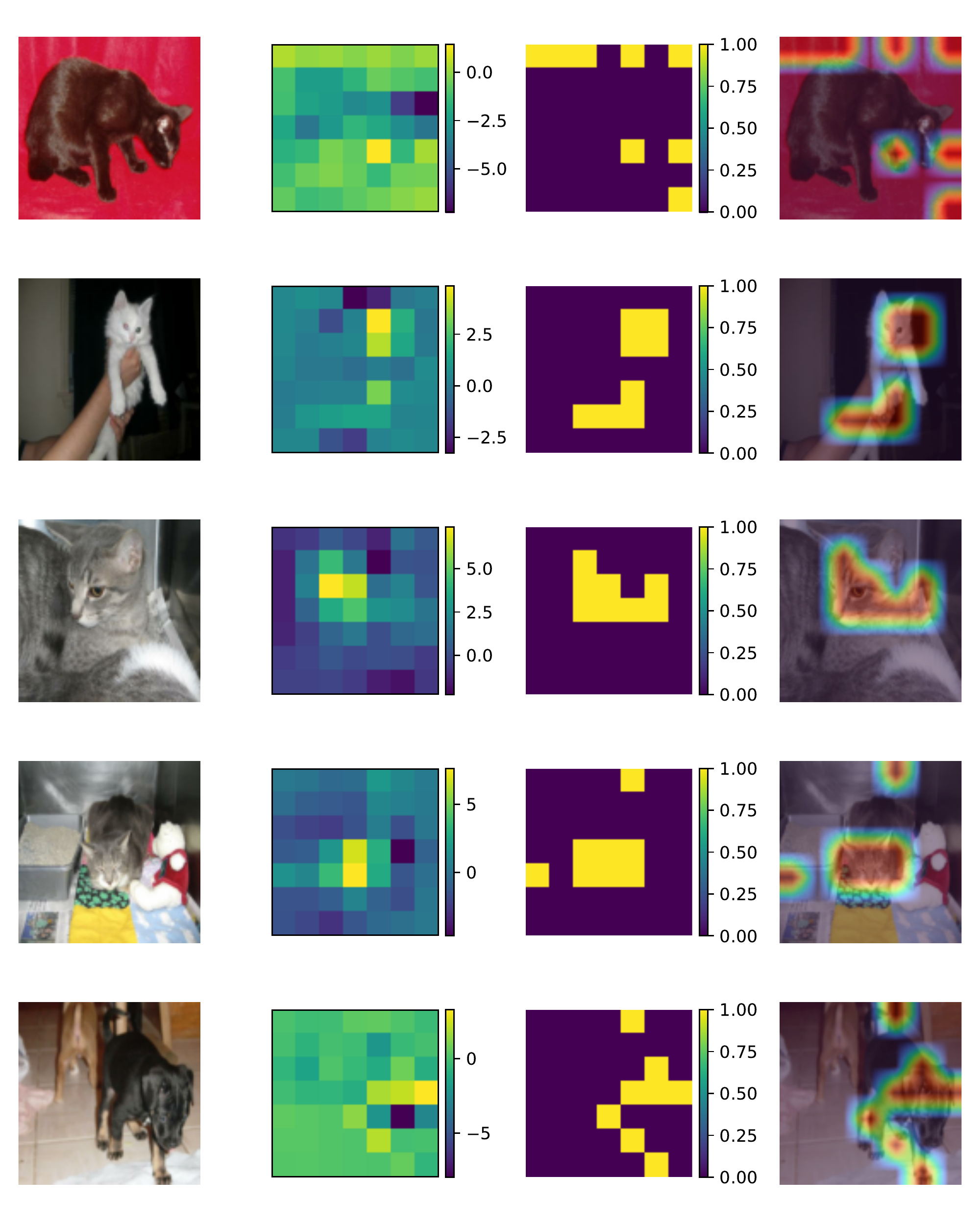}
	\end{center}
	
	\caption{L2X-F -- dataset A}
	\label{fig:first}
\end{figure}

\begin{figure}[h!]
	\begin{center}
		\includegraphics[width=0.5\linewidth]{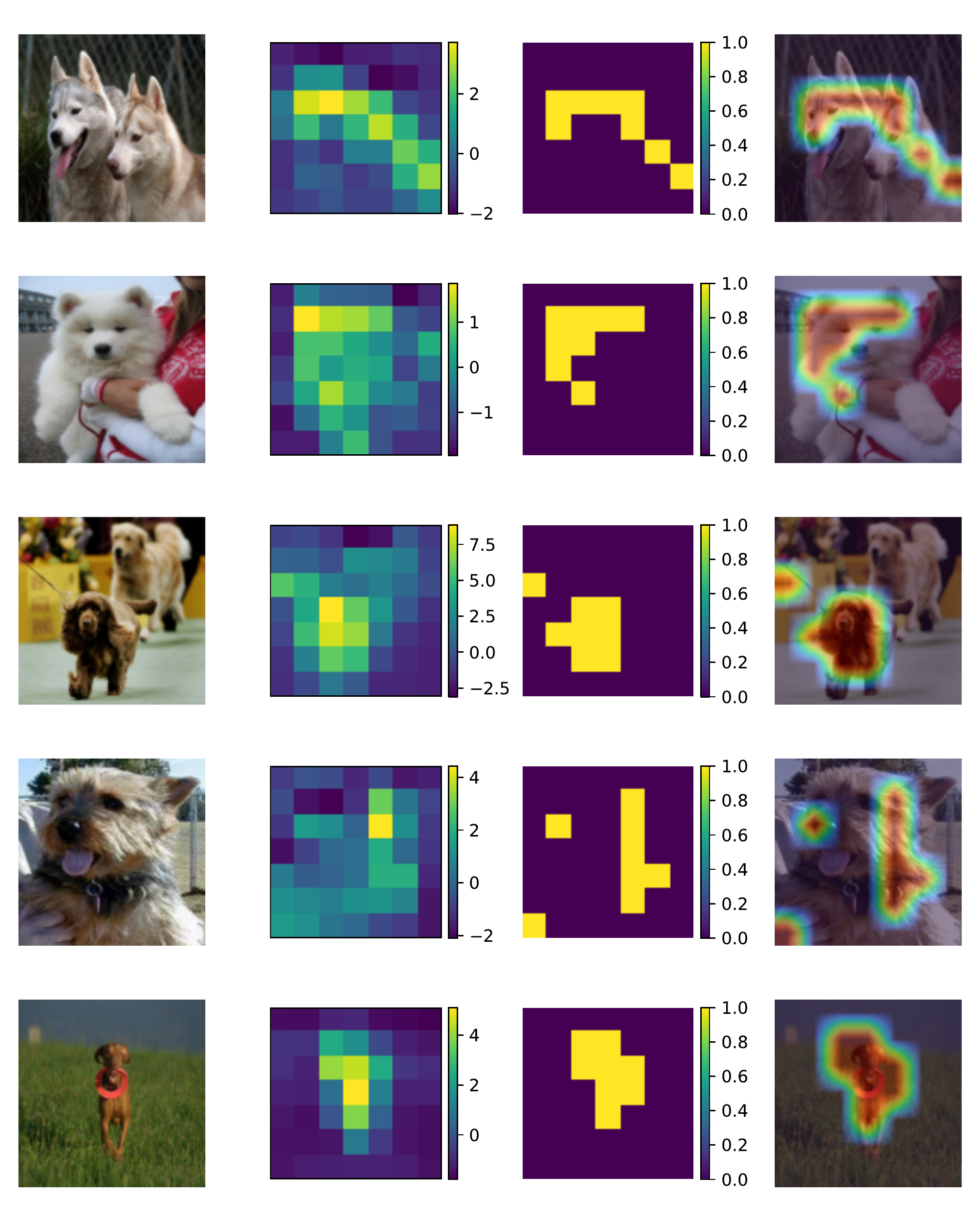}
	\end{center}
	\caption{L2X-F -- dataset B}
\end{figure}

\begin{figure}[h!]
	\begin{center}
		\includegraphics[width=0.5\linewidth]{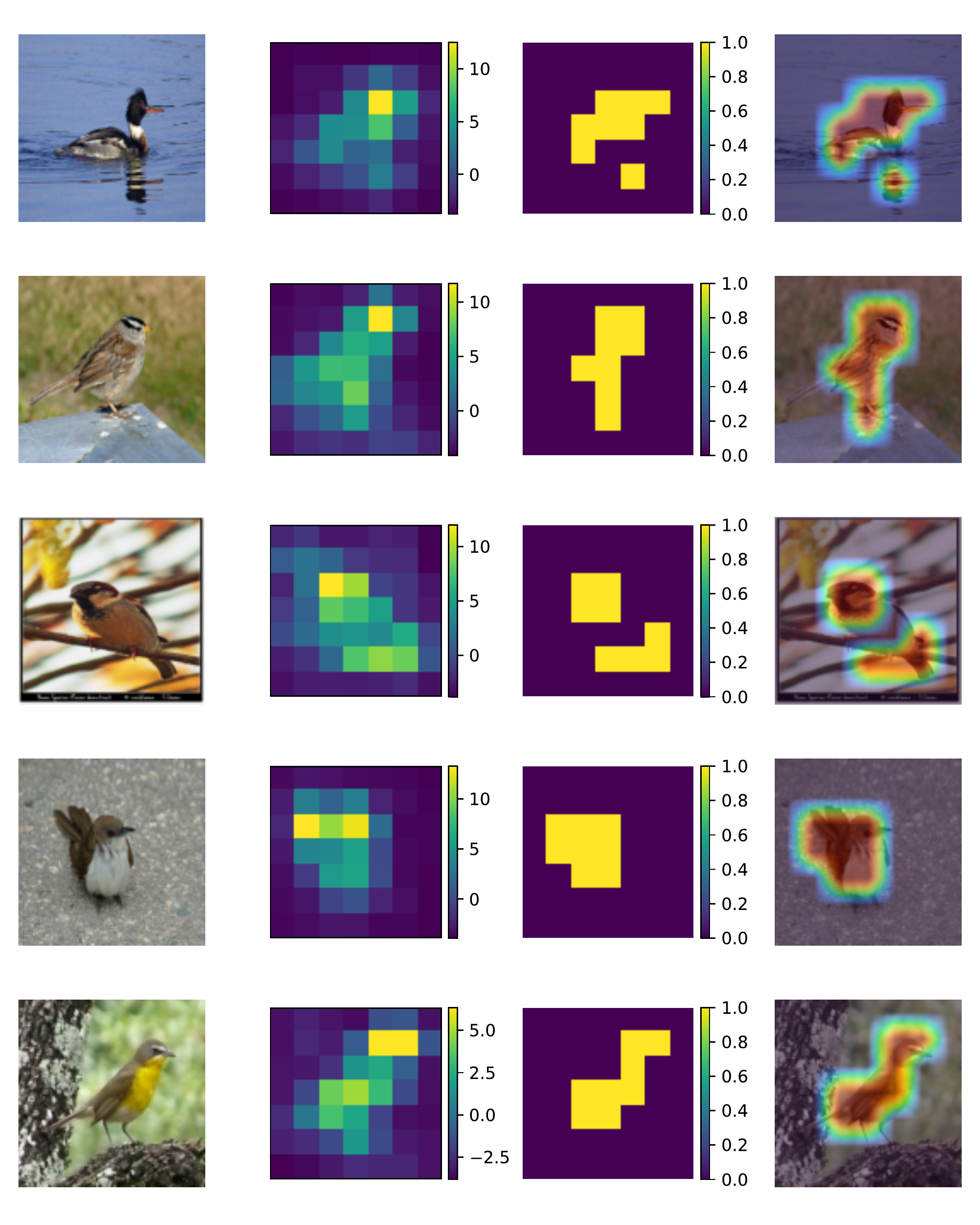}
	\end{center}
	\caption{L2X-F -- dataset C}
\end{figure}

\begin{figure}[h!]
	\begin{center}
		\includegraphics[width=0.5\linewidth]{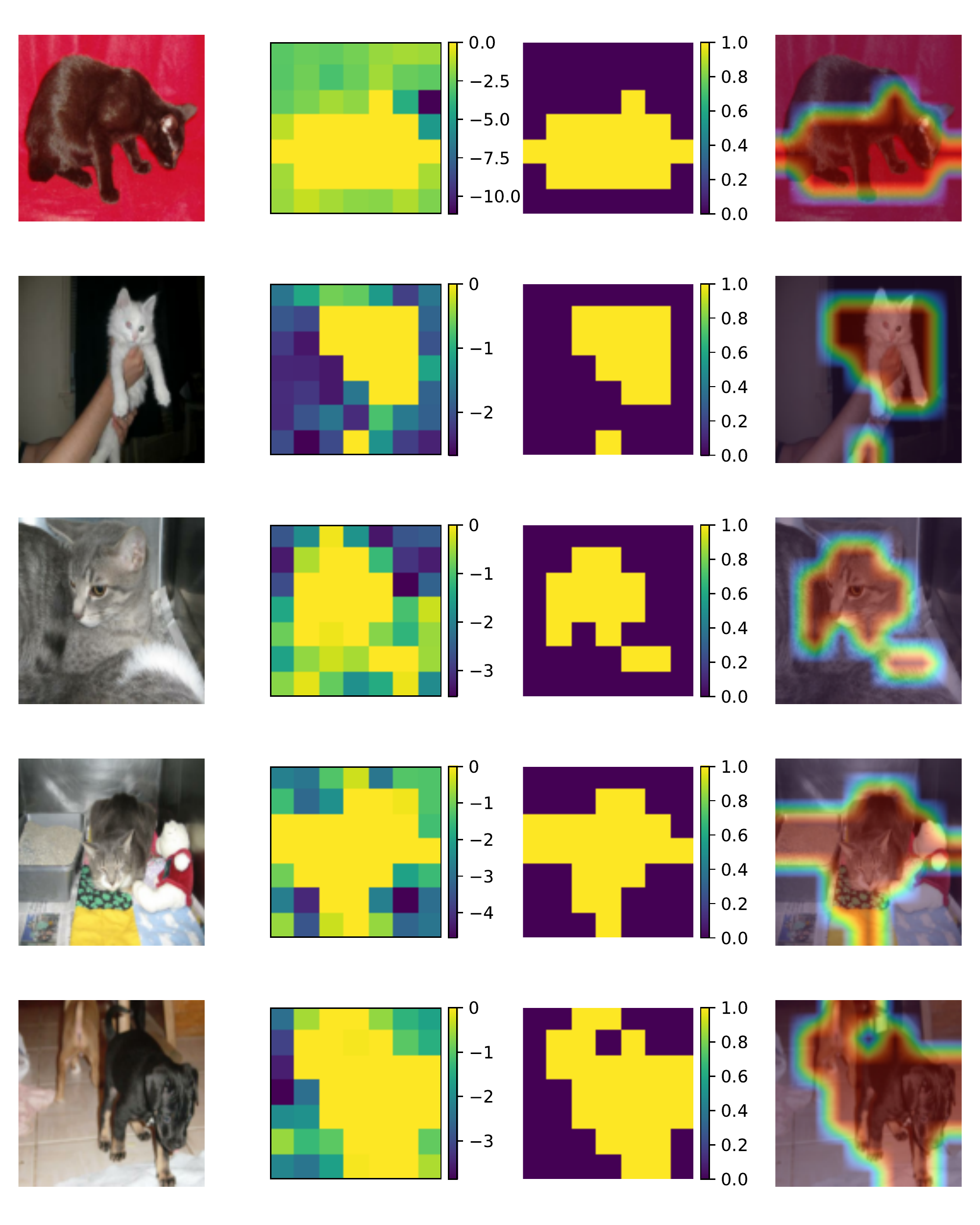}
	\end{center}
	
	\caption{BLA -- dataset A}
\end{figure}

\begin{figure}[h!]
	\begin{center}
		\includegraphics[width=0.5\linewidth]{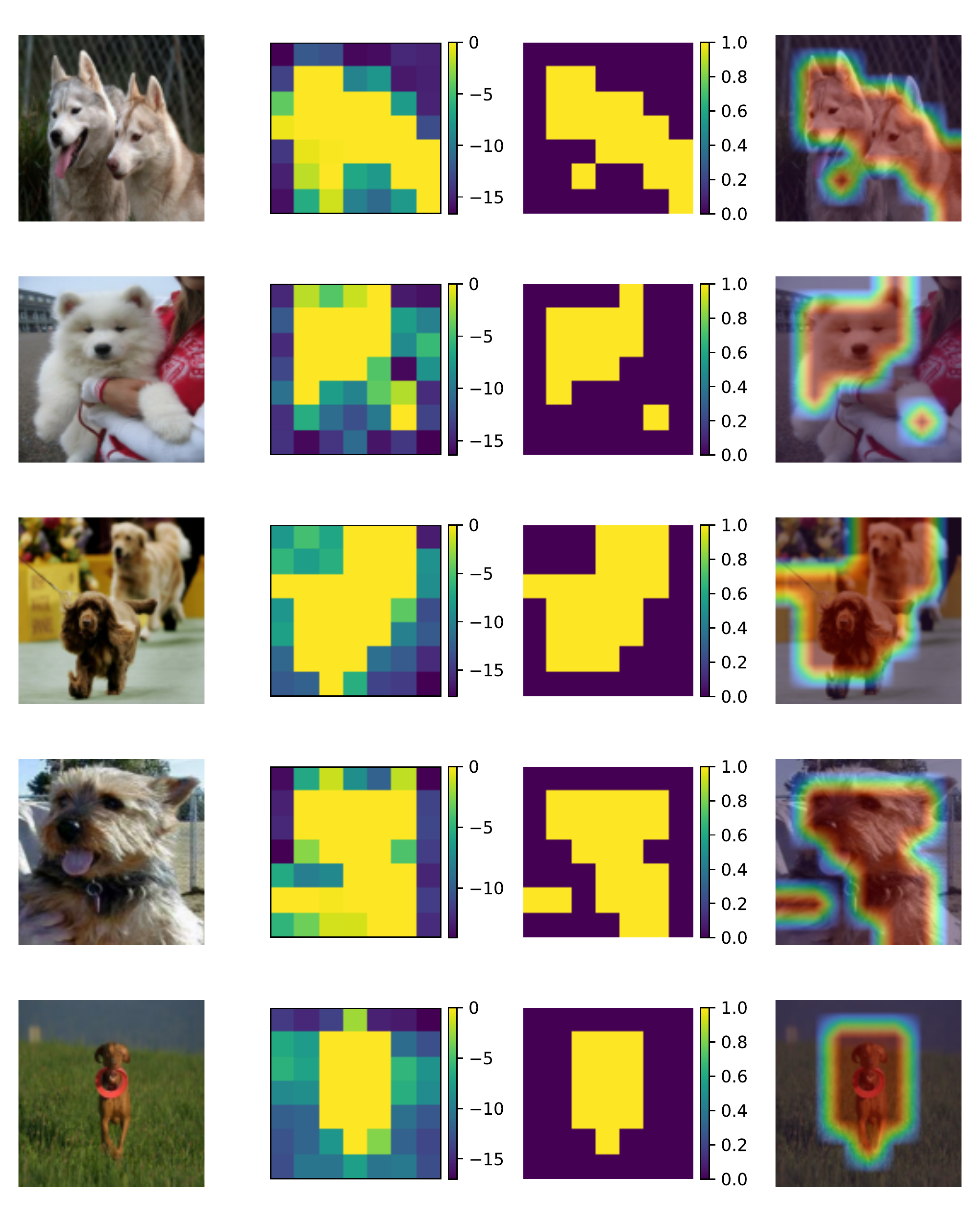}
	\end{center}
	\caption{BLA -- dataset B}
\end{figure}

\begin{figure}[h!]
	\begin{center}
		\includegraphics[width=0.5\linewidth]{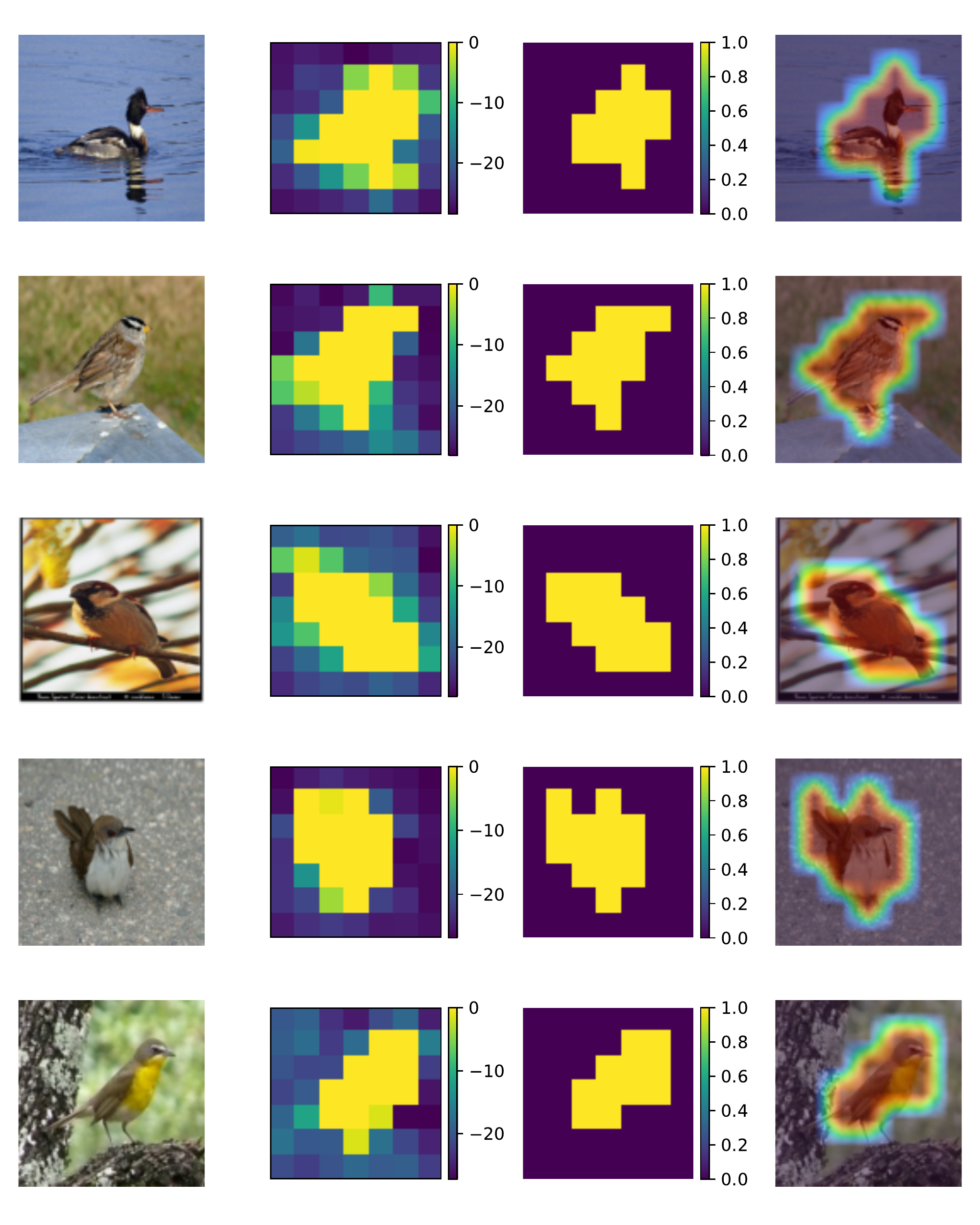}
	\end{center}
	\caption{BLA -- dataset C}
\end{figure}

\begin{figure}[h!]
	\begin{center}
		\includegraphics[width=0.5\linewidth]{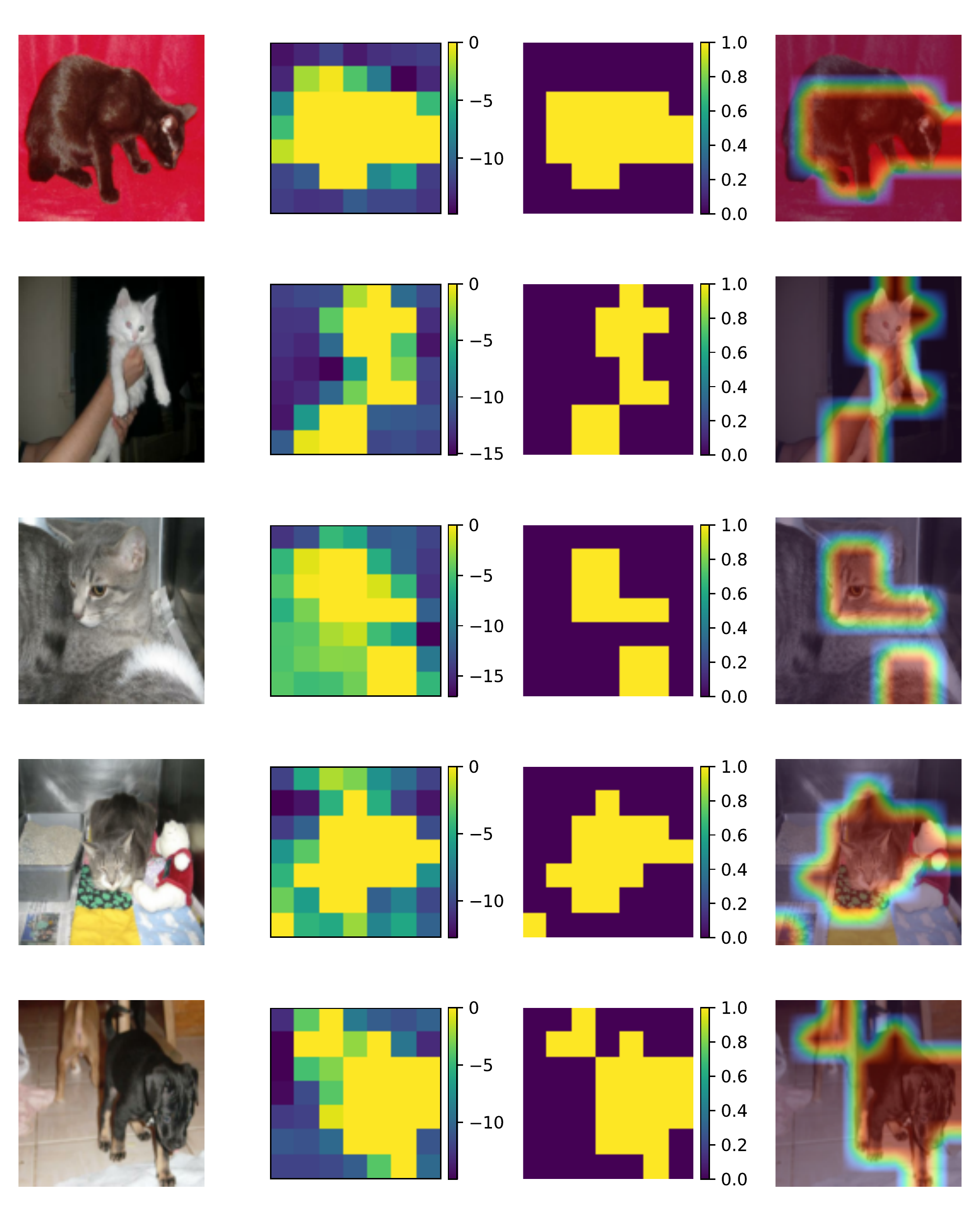}
	\end{center}
	
	\caption{BLA-T -- dataset A}
\end{figure}

\begin{figure}[h!]
	\begin{center}
		\includegraphics[width=0.5\linewidth]{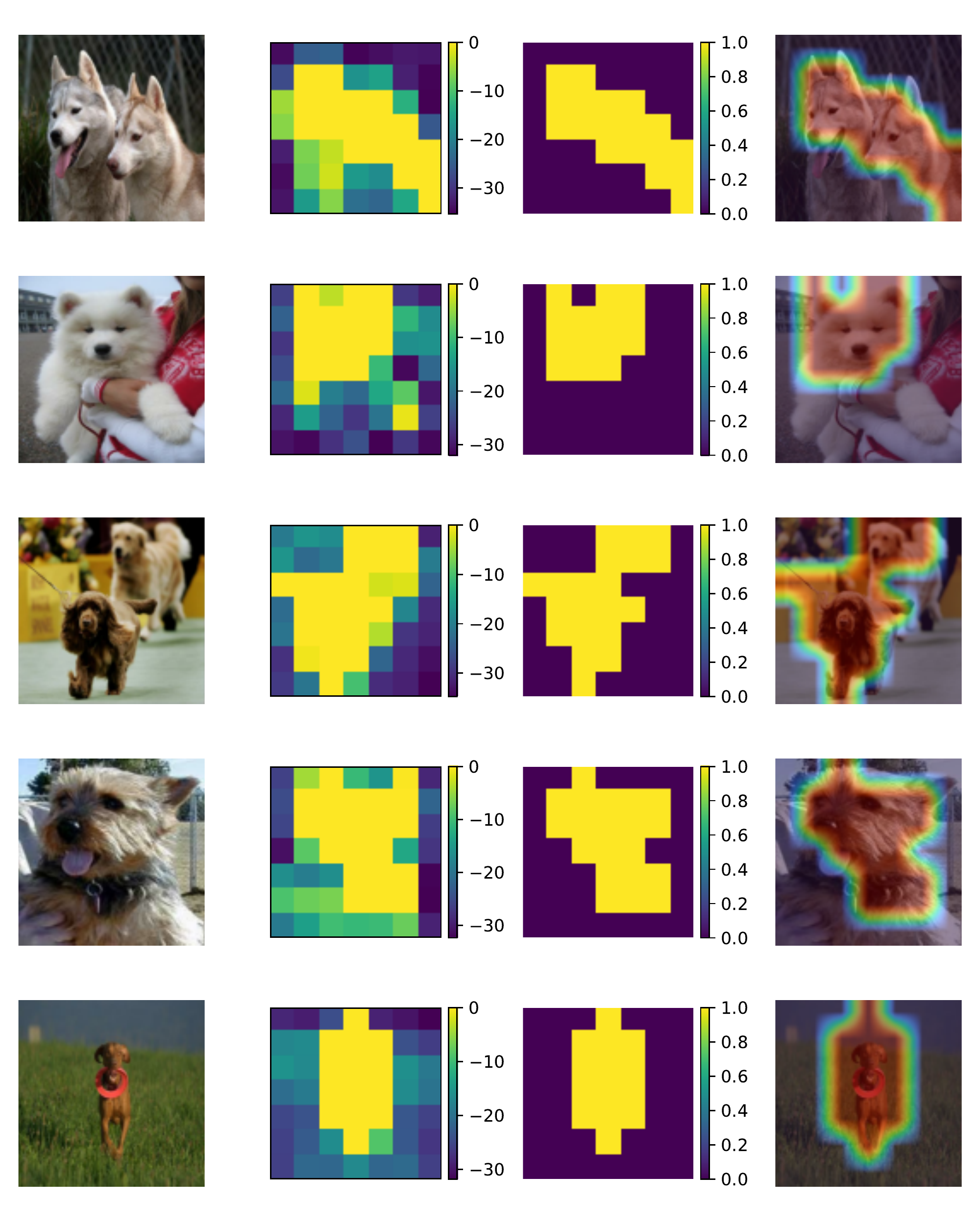}
	\end{center}
	\caption{BLA-T -- dataset B}
\end{figure}

\begin{figure}[h!]
	\begin{center}
		\includegraphics[width=0.5\linewidth]{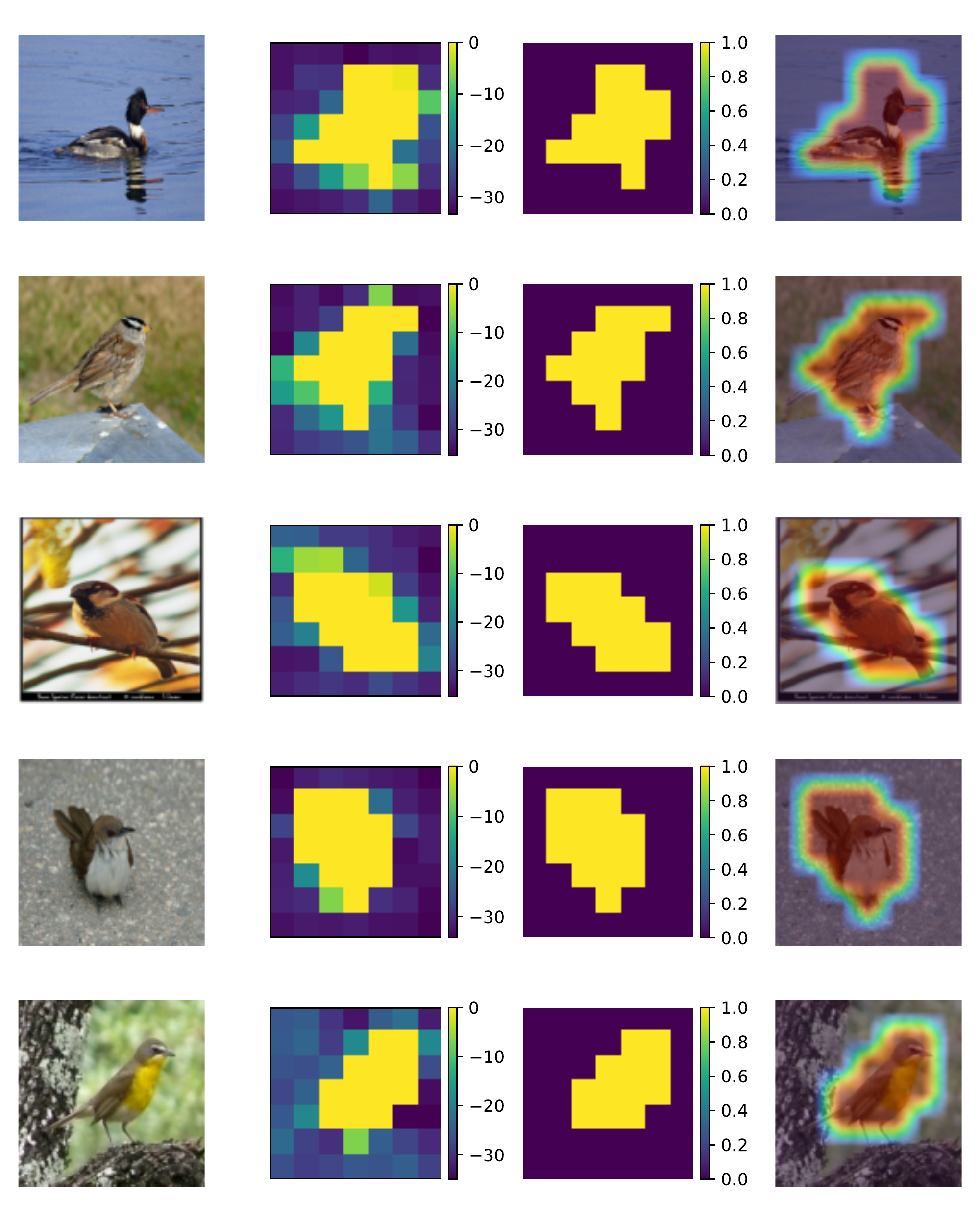}
	\end{center}
	\caption{BLA-T -- dataset C}
\end{figure}

\begin{figure}[h!]
	\begin{center}
		\includegraphics[width=0.5\linewidth]{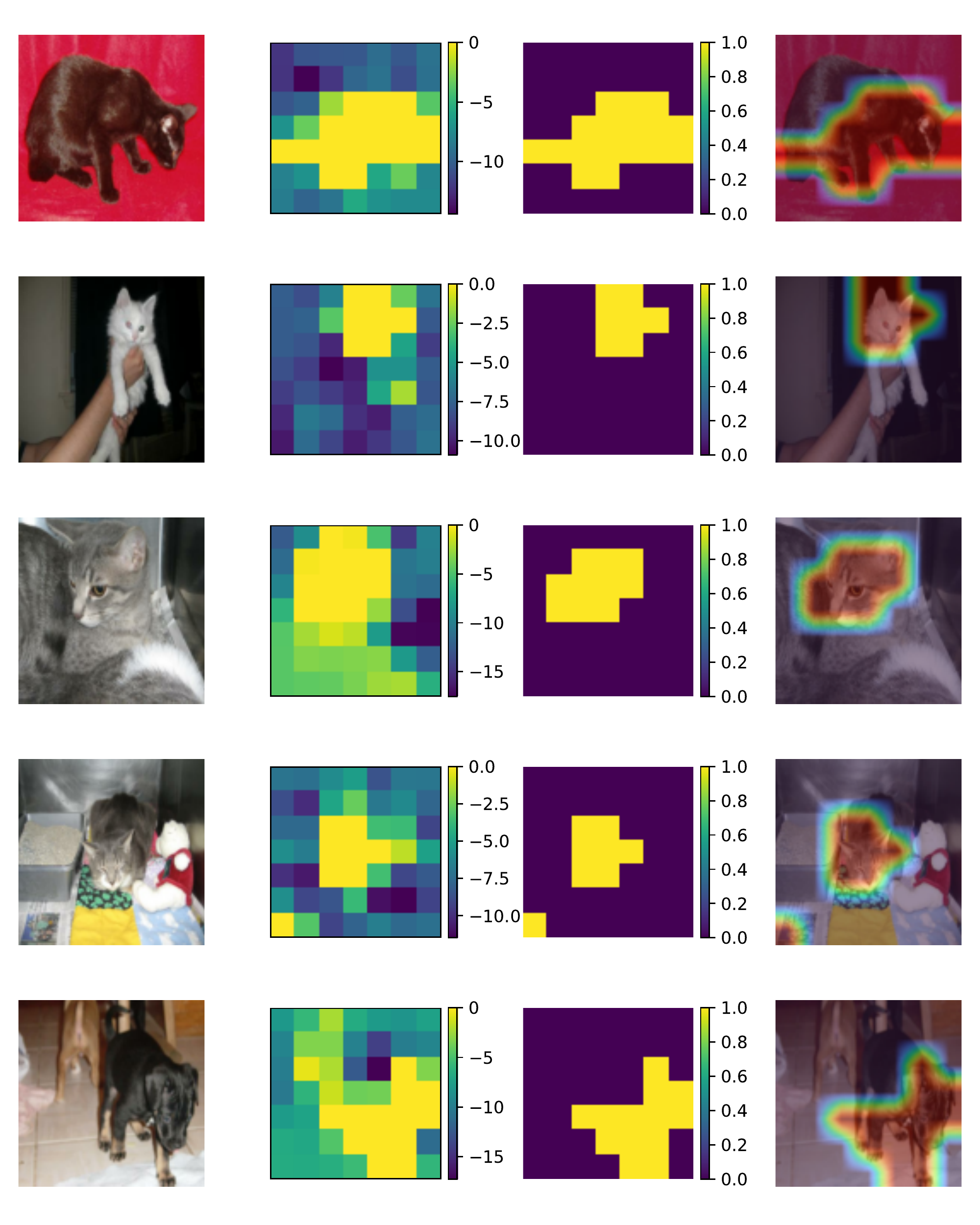}
	\end{center}
	
	\caption{BLA-PH -- dataset A}
\end{figure}

\begin{figure}[h!]
	\begin{center}
		\includegraphics[width=0.5\linewidth]{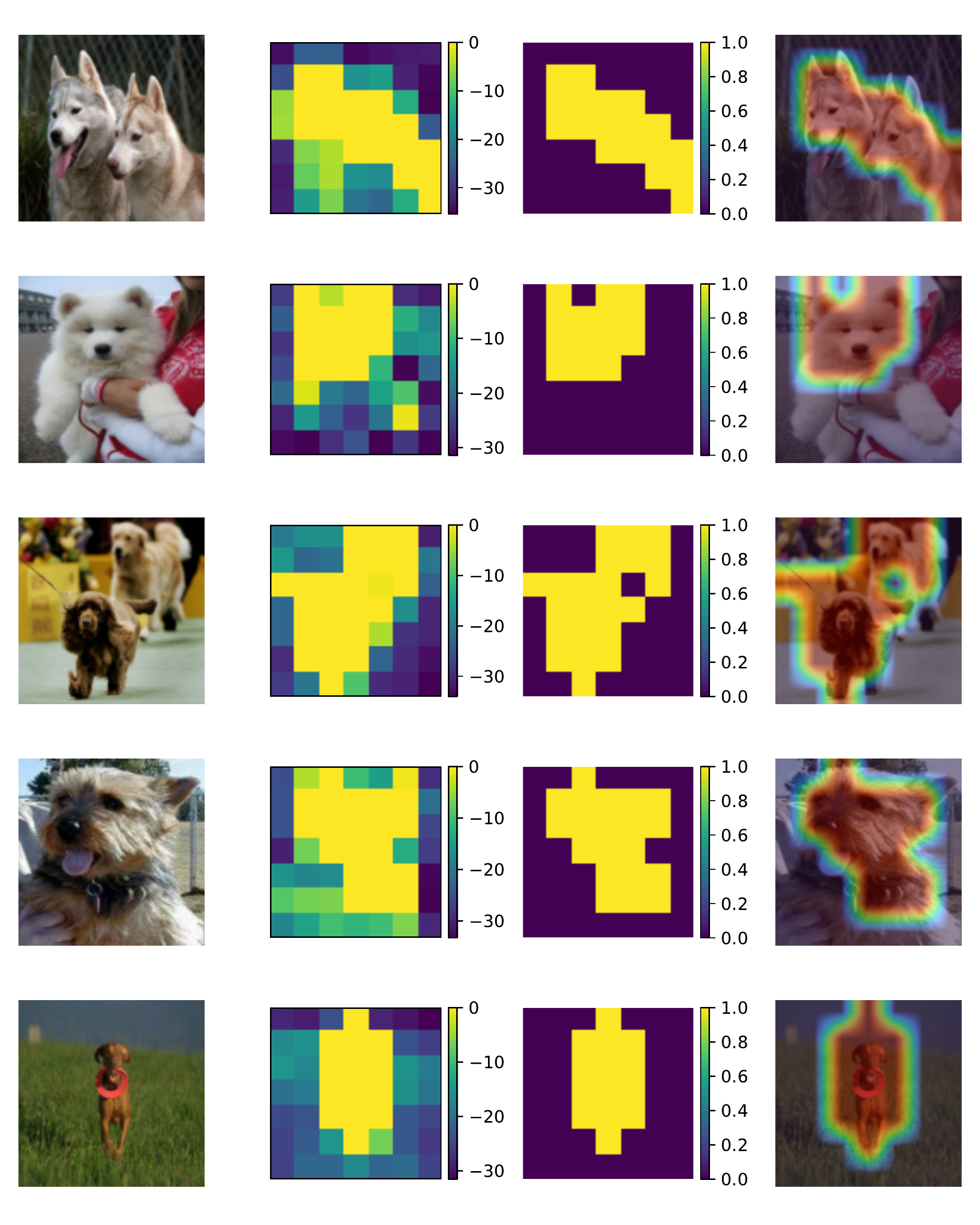}
	\end{center}
	\caption{BLA-PH -- dataset B}
\end{figure}

\begin{figure}[h!]
	\begin{center}
		\includegraphics[width=0.5\linewidth]{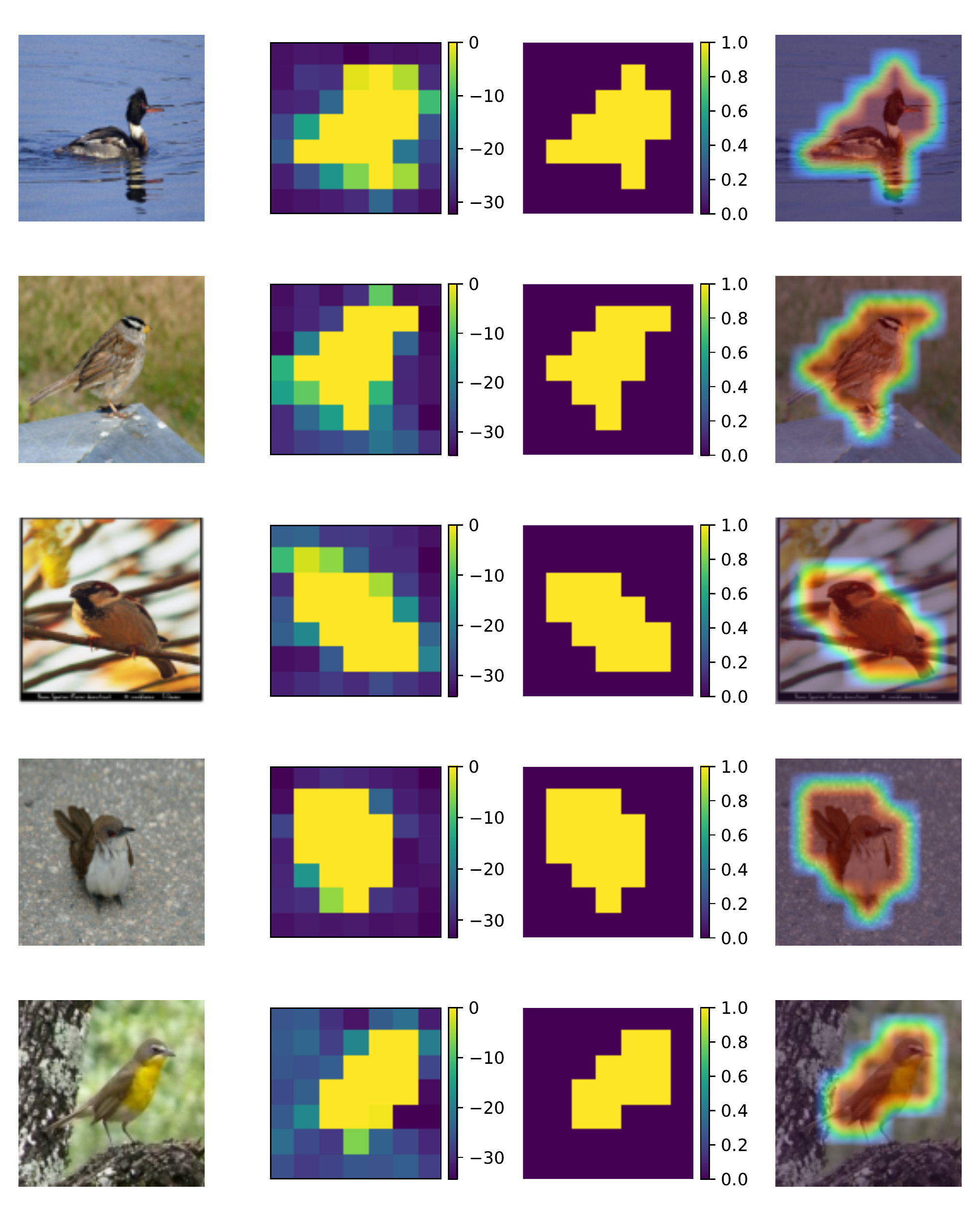}
	\end{center}
	\caption{BLA-PH -- dataset C}
	\label{fig:last}
\end{figure}

\newpage
\clearpage
\section{More context:  attention with global concept vector}

\begin{figure}[h!]
	\begin{center}
		\includegraphics[width=0.6\linewidth]{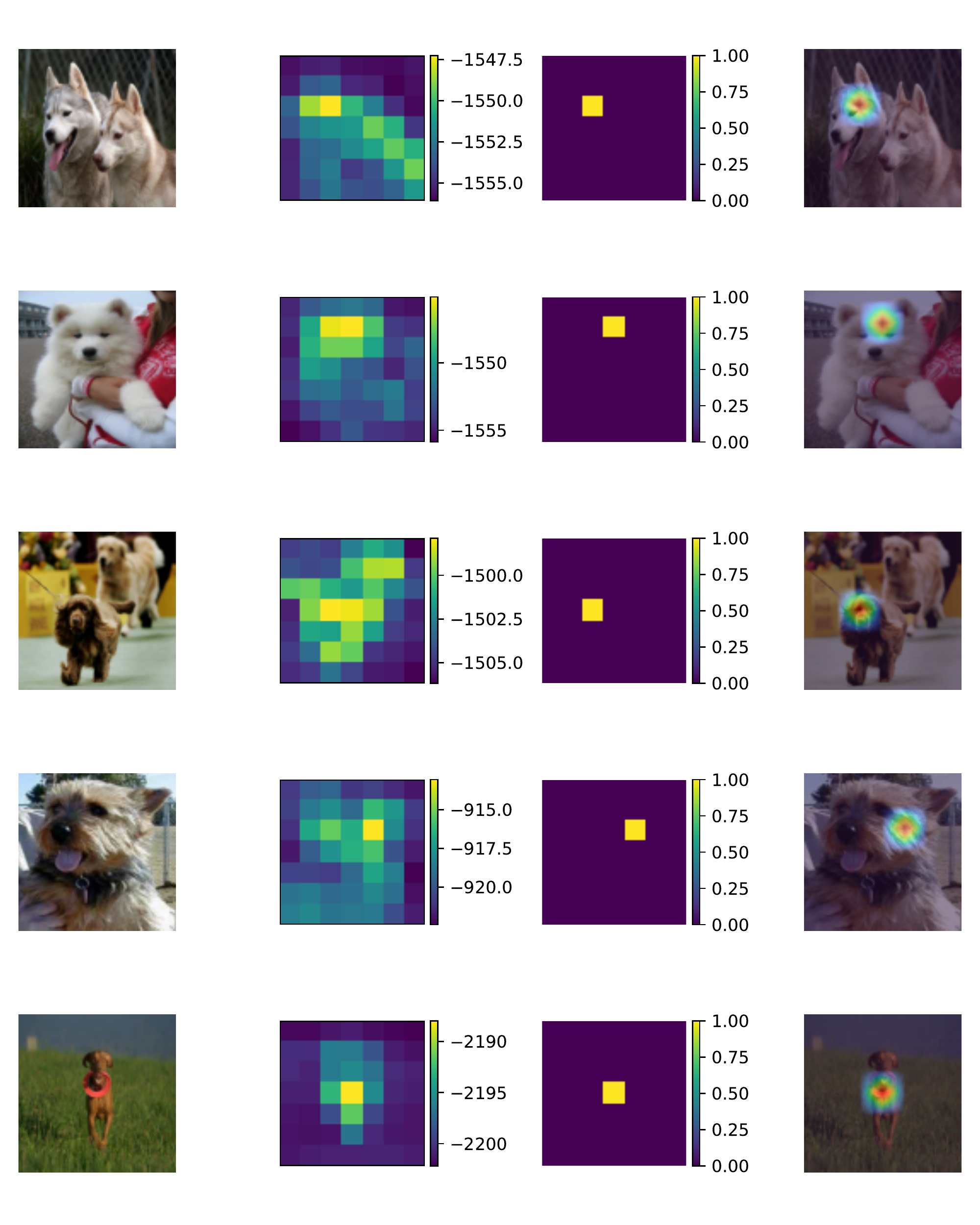}
	\end{center}
	
	\caption{Attempt of using \cite{jetley2018learn} for explainability.}
	\label{fig:global-concept}
\end{figure}

\end{document}